\documentclass{article} 
\usepackage{iclr2027_conference,times}

\usepackage{amsmath,amsfonts,bm}

\def\eqref#1{equation~\ref{#1}}

\def\1{\bm{1}}

\DeclareMathAlphabet{\mathsfit}{\encodingdefault}{\sfdefault}{m}{sl}
\SetMathAlphabet{\mathsfit}{bold}{\encodingdefault}{\sfdefault}{bx}{n}

\usepackage{hyperref}
\usepackage{pifont}
\usepackage{graphicx}
\usepackage{float}
\usepackage{array}
\usepackage{booktabs}
\usepackage{tabularx}
\usepackage{amsmath}
\usepackage{multirow}
\usepackage{caption}
\usepackage[most]{tcolorbox}
\usepackage{CJKutf8}
\usepackage{makecell}
\usepackage{enumitem}
\usepackage{subcaption}
\usepackage[table]{xcolor}
\usepackage{array}
\usepackage[table]{xcolor}
\usepackage{pifont}
\usepackage{placeins}
\usepackage{titlesec}
\usepackage{needspace}
\usepackage{enumitem}
\usepackage{wrapfig}
\usepackage{titletoc}
\usepackage{xurl}
\newtcolorbox{promptbox}[1]{
  colback=gray!5,
  colframe=black,
  boxrule=0.6pt,
  arc=0pt,
  left=6pt,
  right=6pt,
  top=6pt,
  bottom=6pt,
  title=\textbf{#1},
  fonttitle=\normalsize,
  fontupper=\small,
  breakable
}

\newcolumntype{Y}{>{\centering\arraybackslash}X}

\title{C-HAT-Bench: Benchmarking Chinese AI-Text Detection Beyond Fully Generated Text}

\author{
Qing~Yang$^{1}$, Zixiang~Luo$^{1}$,
Zhenyu~Mao$^{1}$, Zezheng~Wu$^{1}$,
Xinghe~Cheng$^{2}$, Haibo~Chen$^{3}$ \\
\textbf{Qinggang Zhang} $^{4}$, \textbf{Jiapu Wang}$^{3}$\thanks{Corresponding author.},
$\ $\textbf{Jingwei Zhang}$^{1}$\\
$^{1}$Guilin University of Electronic Technology, China,\qquad
$^{2}$Jinan Unaiversity, China\\
$^{3}$Nanjing University of Science and Technology, China,\qquad
$^{4}$Jilin Unaiversity, China\\
\texttt{\{gtyqing,gtzjw\}@hotmail.com},\qquad
\texttt{po1aris@mails.guet.edu.cn}\\
\texttt{jnuchengxh@hotmail.com},\qquad
\texttt{jiapu.wang@njust.edu.cn}
}

\iclrfinalcopy
\begin{document}

\maketitle

\lhead{Preprint}

\begin{abstract}
Large Language Models (LLMs) increasingly participate in writing by modifying or extending human drafts, causing machine involvement to vary in both form and extent. Yet most Machine-Generated Text (MGT) detectors are evaluated only on fully human-written versus fully AI-generated text. Because human--AI collaboration can weaken or redistribute cues associated with machine generation, strong performance under this binary setting may overstate detector reliability. This mismatch remains underexplored in Chinese: detection cues are shaped by tokenization and language-specific text distributions, yet controlled resources spanning production settings, domains, and generators remain limited. To fill this gap, we present a \textbf{\underline{C}}hinese \textbf{\underline{H}}uman-\textbf{\underline{A}}I Collaborative \textbf{\underline{T}}ext Detection \textbf{\underline{Bench}}mark (\textbf{C-HAT-Bench}), a unified benchmark that links $5,000$ human-written source texts from five domains to more than $240,000$ variants produced using six generative models under \textit{Prefix-Conditioned Continuation} as a reference setting and three collaborative production modes. We evaluate $21$ detectors through four protocols spanning zero-shot and pretrained supervised document-level detection, boundary localization, and cross-condition generalization. Relative to Prefix-Conditioned Continuation, mean AUROC across document-level detectors is $12.0\%$ lower on the collaborative production modes, with the largest detector-specific relative decrease reaching $44.4\%$. Transfer across collaborative production modes is also asymmetric, indicating that performance in a given production setting is not a reliable predictor of performance in other production settings.\footnote{Code and data are available at:
\href{https://anonymous.4open.science/r/anonymous-submission-D407/}
{https://anonymous.4open.science}.}
\end{abstract}

\section{Introduction}
Recent advances in Large Language Models (LLMs) \citep{pan2024unifying,zhao2026survey} have substantially enhanced text-generation capabilities, contributing to their broader adoption in writing and information production. As LLMs are increasingly integrated into writing workflows, the resulting texts may reflect contributions from both human authors and generative models, making the extent of machine involvement difficult to determine. Reliable machine-generated text (MGT) detection is therefore needed to support provenance assessment, particularly when generated content is used to disseminate misinformation or facilitate academic misconduct \citep{brown2020language,wang2024m4gtbench,wu2025survey,quaremba2026tsmbench}.



Yet reliable MGT detection remains challenging because the process through which a text is produced is not directly observable, while the cues associated with machine generation depend on the generator and its role in writing. Prevailing benchmarks still emphasize binary discrimination between fully human-written and fully AI-generated text, typically using outputs elicited by generic prompts \citep{wu2025survey,wang2024m4gtbench,quaremba2026tsmbench}. Although this setting provides a useful controlled baseline, it does not capture human-AI collaborative writing, in which machine involvement varies in form and extent. Consequently, performance measured on fully AI-generated text does not, by itself, establish detector reliability under human-AI collaboration \citep{wang2024m4gtbench,quaremba2026tsmbench}.

Moreover, process-aware MGT evaluation in Chinese must account for detection cues that are sensitive to tokenization and language-specific text distributions. Chinese lacks explicit word boundaries, and its writing conventions vary substantially across domains \citep{tsang2025chinese,qing2026cred}. Yet existing Chinese benchmarks do not enable controlled comparisons of human--AI collaboration modes across domains and generators \citep{guo2023hc3,macko2023multitude,wang2024m4,qing2026cred}. A unified benchmark is therefore needed to assess detector reliability across human--AI writing processes in Chinese.

\begin{figure}[t]
    \centering
    \includegraphics[width=1.0\textwidth]{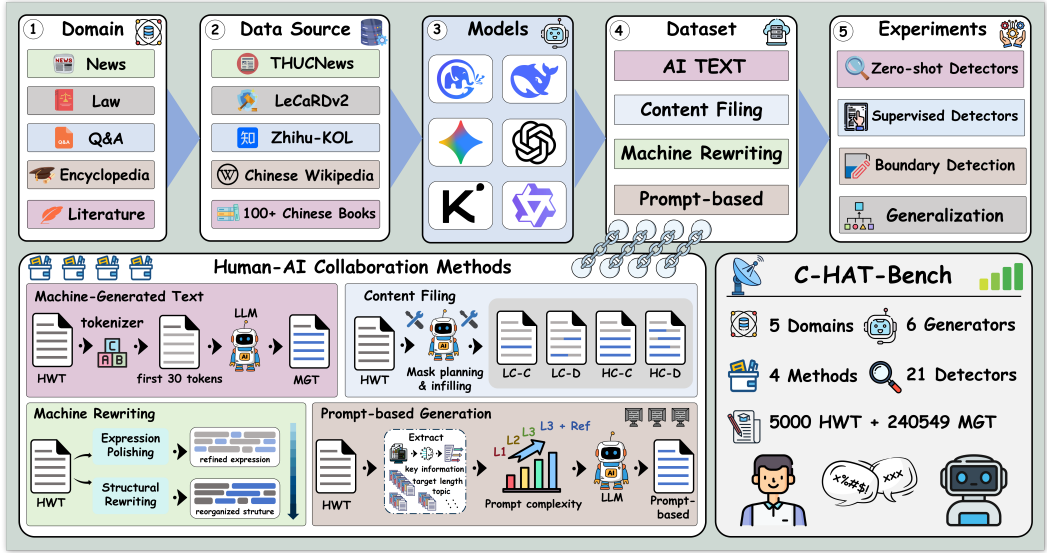}
    \caption{Overview of the C-HAT-Bench framework.}
    \label{fig:chact-bench-overview}
    \vspace{-10pt}
\end{figure}

To enable such evaluation, we construct the Chinese Human--AI Collaborative Text Dataset (\textbf{C-HAT-240K}), which links each human-written source text to multiple machine-involved variants. C-HAT-240K comprises $5,000$ source texts from five domains---news, law, question answering, encyclopedic writing, and literature---and $240,539$ variants produced with six LLMs. As shown in Figure~\ref{fig:chact-bench-overview}, the construction includes Prefix-Conditioned Continuation as a reference setting and three collaborative production modes: Content Filling, Machine Rewriting, and Prompt-based Generation. The reference setting retains a human-written prefix, whereas the collaborative modes vary generated-span coverage and placement, rewriting intensity, and prompt guidance. This source-linked design enables controlled comparisons of detector performance across production processes.


Building on C-HAT-240K, we introduce a \textbf{\underline{C}}hinese \textbf{\underline{H}}uman-\textbf{\underline{A}}I Collaborative \textbf{\underline{T}}ext Detection \textbf{\underline{Bench}}mark (\textbf{C-HAT-Bench}), a unified framework for evaluating $21$ detectors under four protocols. The document-level evaluation comprises separate protocols for $10$ zero-shot detectors and $7$ pretrained supervised detectors across all production settings. A third protocol assesses $2$ detectors on the Content Filling subset for localizing boundaries between human-written and machine-generated content. The fourth examines cross-condition generalization by training $2$ supervised detectors on one condition at a time and evaluating their transfer under changes in domain, production setting, or generator. The main contributions of this work are threefold:

\begin{itemize}[
    leftmargin=1.2em,
    labelindent=0pt,
    labelwidth=0.7em,
    labelsep=0.5em,
    itemsep=2pt,
    topsep=2pt
]

\item \textbf{Dataset.}
We construct C-HAT-240K, a Chinese dataset that links 5,000 human-written source texts from five domains to more than 240,000 variants produced using six generative models. The variants cover one reference setting and three collaborative production modes;
\item \textbf{Unified evaluation benchmark.}
We introduce \textbf{C-HAT-Bench}, a unified framework for evaluating $21$ detectors through four protocols. These comprise separate document-level protocols for ten zero-shot and seven pretrained supervised detectors, a boundary-localization protocol with two detectors, and a cross-condition generalization protocol with two supervised detectors;
\item \textbf{Multi-dimensional analysis.}
We analyze detector sensitivity to the proportion and placement of machine-generated content, the degree of model intervention, and the level of guidance provided by prompts. The analysis further examines boundary localization and transfer across production settings, domains, and generators.
\end{itemize}

\noindent\textbf{Key findings.}
Relative to Prefix-Conditioned Continuation, mean AUROC across detectors is $12.0\%$ lower on the three collaborative production modes, with the largest detector-specific relative decrease reaching $44.4\%$. Boundary matching improves under dispersed filling, while transfer across collaborative production modes is asymmetric, indicating that performance under one production setting does not reliably predict performance under another.

\section{Related Work}

\noindent\textbf{AIGC Text Detection Methods}\quad
AIGC text detection has progressed from document-level source classification to fine-grained attribution in human--AI collaborative text. Early work focused on statistical traces in generated text: \citet{gehrmann2019gltr} visualized token probabilities and ranks, while \citet{verma2024ghostbuster} learned classifiers from probability-derived features. As reliance on labeled data became a central limitation, \citet{mitchell2023detectgpt} introduced probability-curvature-based detection, followed by log-rank, conditional-curvature, perplexity-based, and query-based methods~\citep{su2023detectllm,bao2024fastdetectgpt,hans2024binoculars,zhu2023beatllms}. A complementary line embeds detectable signals during generation: \citet{kirchenbauer2023watermark} proposed a statistical watermark for language models, and \citet{dathathri2024scalable} extended watermarking to scalable identification of model outputs. Subsequent studies addressed robustness under domain shifts, adversarial manipulation, and machine rewriting~\citep{bhattacharjee2023conda,hu2023radar,kuznetsov2024restricted,chen2025imitate,hao2025rewrite}. With the emergence of mixed-origin text, \citet{lee2022coauthor} introduced a human--AI collaborative writing dataset, while later studies investigated machine revision, sentence-level attribution, boundary segmentation, and generated-span localization~\citep{zhang2024coauthor,wang2023seqxgpt,zeng2024sentences,jiang2025sendetex,li2026segmenting,tolstykh-etal-2026-gigacheck}. These developments move detection beyond document-level authorship, but realistic collaboration processes remain insufficiently covered.

\noindent\textbf{Chinese AIGC Corpora}\quad
Chinese NLP research first established broad linguistic resources for different levels of analysis. \citet{xue2005pennctb} introduced the Penn Chinese Treebank for syntactic annotation, while \citet{zhou2014cdtb} developed the Chinese Discourse TreeBank for discourse-relation analysis. Large-scale pretraining resources were later provided by \citet{yuan2021wudaocorpora}. Task-oriented datasets further expanded domain coverage, including legal retrieval through LeCaRDv2~\citep{li2024lecardv2}. For AIGC detection, \citet{guo2023hc3} constructed a Chinese question-answering corpus pairing human and ChatGPT responses, while MULTITuDE and M4 broadened the coverage of languages, domains, and generators~\citep{macko2023multitude,wang2024m4}. C-ReD further introduced Chinese detection data derived from real-world-inspired prompts across domains and models~\citep{qing2026cred}. These resources provide important foundations, but they do not jointly control collaboration modes, machine-involvement levels, and source-linked local authorship.

\noindent\textbf{MGT Detection Benchmarks}\quad
MGT benchmarks have consequently moved from source identification toward evaluation under changing data conditions. \citet{uchendu2021turingbench} established document-level detection and generator attribution, after which MAGE and MGTBench broadened comparisons across domains, generators, and detection methods~\citep{li2024mage,he2024mgtbench}. Robustness benchmarks such as BUST, RAID, and DetectRL then examined detector behavior under perturbations, adversarial attacks, evasion settings, and real-world generation conditions~\citep{bust2024benchmark,dugan2024raid,wu2024detectrl}. For mixed-origin text, M4GT-Bench modeled a human-written prefix followed by machine-generated continuation, whereas TSM-BENCH evaluated detection in practical Wikipedia editing tasks~\citep{wang2024m4gtbench,quaremba2026tsmbench}. However, existing benchmarks still lack a unified Chinese setting that connects controlled collaboration construction with local boundary localization and cross-condition generalization.

\section{C-HAT-Bench}
\label{sec:benchmark}

\subsection{Dataset Construction}
\label{sec:benchmark-datasets}

\noindent\textbf{Data Collection.}
C-HAT-240K covers five domains, with human-written source texts for each domain drawn from a distinct dataset. News texts are drawn from
THUCNews\footnote{Dataset available at
\url{http://thuctc.thunlp.org}.}. The Law, Q\&A, and Encyclopedia texts come from
LeCaRDv2~\citep{li2024lecardv2},
Zhihu-KOL\footnote{Dataset available at
\url{https://huggingface.co/datasets/wanqiu6/Zhihu-KOL}.},
and the Chinese Wikipedia corpus on
ModelScope\footnote{Dataset available at
\url{https://www.modelscope.cn/datasets/caolong/zhwiki}.},
respectively. Literature texts come from a locally curated collection
of 102 modern Chinese literary works. Detailed data extraction procedures and dataset details are provided in Appendix~\ref{app:hwt-collection}.

\noindent\textbf{Data Processing.}
Before C-HAT-240K construction, we extract and clean the human-written texts.
For domains other than Literature, we target 1{,}000 texts per domain by
starting from a central text-length target measured in characters,
progressively expanding the selection range toward shorter and longer texts,
and relaxing the Chinese-character proportion threshold when necessary
without falling below a preset minimum. The Literature domain is processed
separately by segmenting works at paragraph and chapter boundaries, removing
non-content material, and retaining 1{,}000 continuous passages. All selected
texts are then normalized, stripped of metadata artifacts and malformed
fragments, and deduplicated; detailed procedures are provided in
Appendix~\ref{app:data-cleaning}.

\noindent\textbf{Dataset Building.} We organize the machine-involved variants into four construction types, each derived from a human-written source and instantiated with the six generative models. Let $x$ and $y$ denote the source and constructed texts, respectively. $\mathcal{M}_{\theta}(u,p)$ denotes a generative model with parameters $\theta$ that produces text from input $u$ under prompt $p$, and $\oplus$ denotes string concatenation.
Detailed descriptions of the generative models are provided in Appendix~\ref{app:generative-models}, while detailed construction rules and prompt templates are given in Appendix~\ref{app:prompt-templates}.

\begin{itemize}[
    label=\textbullet,
    leftmargin=1em,
    labelwidth=0.5em,
    labelsep=0.5em,
    itemindent=0pt,
    align=left,
    itemsep=4pt,
    topsep=6pt,
    parsep=0pt,
    partopsep=0pt
]

\item \textbf{Type I: Machine-Generated Text.}
We retain the first 30 tokens of each source text under the Qwen2.5-3B-Instruct tokenizer and use them as the context for continuation:
\begin{equation}
\label{eq:chact-prefix-construction}
y_{\theta}^{\mathrm{MGT}}(x)=\operatorname{Pref}_{30}^{\tau}(x)\oplus\mathcal{M}_{\theta}\!\left(\operatorname{Pref}_{30}^{\tau}(x),p_{\mathrm{cont}}\right).
\end{equation}
Here, $\mathrm{Pref}_{30}^{\tau}(x)$ is the retained human-written prefix,
$p_{\mathrm{cont}}$ requests a model continuation.  In the following experiments, MGT refers to this prefix-conditioned reference condition.

\item \textbf{Type II: Content Filling.}
We replace selected source spans with generated text while preserving the remaining human-written content. The four conditions combine low or high coverage with contiguous or dispersed spans: \underline{\textit{1. LC-C:}} low coverage in one contiguous suffix; \underline{\textit{2. LC-D:}} low coverage in two or three dispersed spans; \underline{\textit{3. HC-C:}} high coverage in one contiguous suffix; and \underline{\textit{4. HC-D:}} high coverage in three or four dispersed spans. Low and high coverage correspond to 15--35\% and 50--75\% of the source characters, respectively:
\begin{equation}
\label{eq:chact-infill-construction}
y_{\theta,t}^{\mathrm{CF}}(x)=h_{t,0}\oplus\bigoplus_{j=1}^{K_t}\left[\mathcal{M}_{\theta}(\tilde{x}_{t,j},p_{t,j})\oplus h_{t,j}\right],\quad t\in\mathcal{T}_{\mathrm{CF}}.
\end{equation}
Here, $\mathcal{T}_{\mathrm{CF}}$ contains the four conditions, $h_{t,j}$ denotes an unchanged source segment, and $\tilde{x}_{t,j}$ is the masked source after prior replacements are inserted. The value of $K_t$ and span layout are determined by condition $t$, while $p_{t,j}$ specifies the target span and its length requirement.

\item \textbf{Type III: Machine Rewriting.}
We provide the complete source text to the model under two rewriting settings:
\begin{equation}
\label{eq:chact-rewrite-construction}
y_{\theta,c}^{\mathrm{MR}}(x)=\mathcal{M}_{\theta}(x,p_c),\quad c\in\{\mathrm{LR},\mathrm{DR}\}.
\end{equation}
Here, \underline{\textit{LR}} (Light Rewriting) performs local wording and fluency adjustments while retaining the original organization; \underline{\textit{DR}} (Deep Rewriting) allows broader changes to expression and organization while preserving the source topic, main facts, and domain style.

\item \textbf{Type IV: Prompt-based Generation.}
GPT-5.5 first extracts a shared structured representation from each source, after which level-specific fields and writing instructions are assembled into the generation prompt:
\begin{equation}
\label{eq:chact-prompt-construction}
y_{\theta,\ell}^{\mathrm{PG}}(x)=\mathcal{M}_{\theta}\!\left(F_{\ell}\!\left(E_{\mathrm{GPT\text{-}5.5}}(x)\right)\oplus r_{\ell},p_{\ell}\right),\quad \ell\in\{\mathrm{L1},\mathrm{L2},\mathrm{L3},\mathrm{L3+Ref}\}.
\end{equation}
Here, $E_{\mathrm{GPT\text{-}5.5}}$ extracts the shared structured fields,
$F_{\ell}$ selects and formats the fields used at level $\ell$, $p_{\ell}$
specifies the corresponding writing instructions, and $r_{\ell}$ denotes an
optional human-written reference from the same domain. For the first three
levels, $r_{\ell}$ is empty.
\emph{\underline{L1}} uses task information only;
\emph{\underline{L2}} adds the key content to the task information;
\emph{\underline{L3}} further incorporates domain-specific constraints; and
\emph{\underline{L3+Ref}} retains the L3 configuration while additionally
providing a different human-written text from the same domain as $r_{\ell}$.

\end{itemize}

\noindent\textbf{Dataset Quality Improvement.}
After constructing the four machine-involved subsets, we first retain only samples whose generated-to-source character-length ratio falls within $[0.6,1.4]$. We 
\begin{wrapfigure}{r}{0.40\textwidth}
    \vspace{-10pt}
    \centering
    \includegraphics[width=\linewidth]{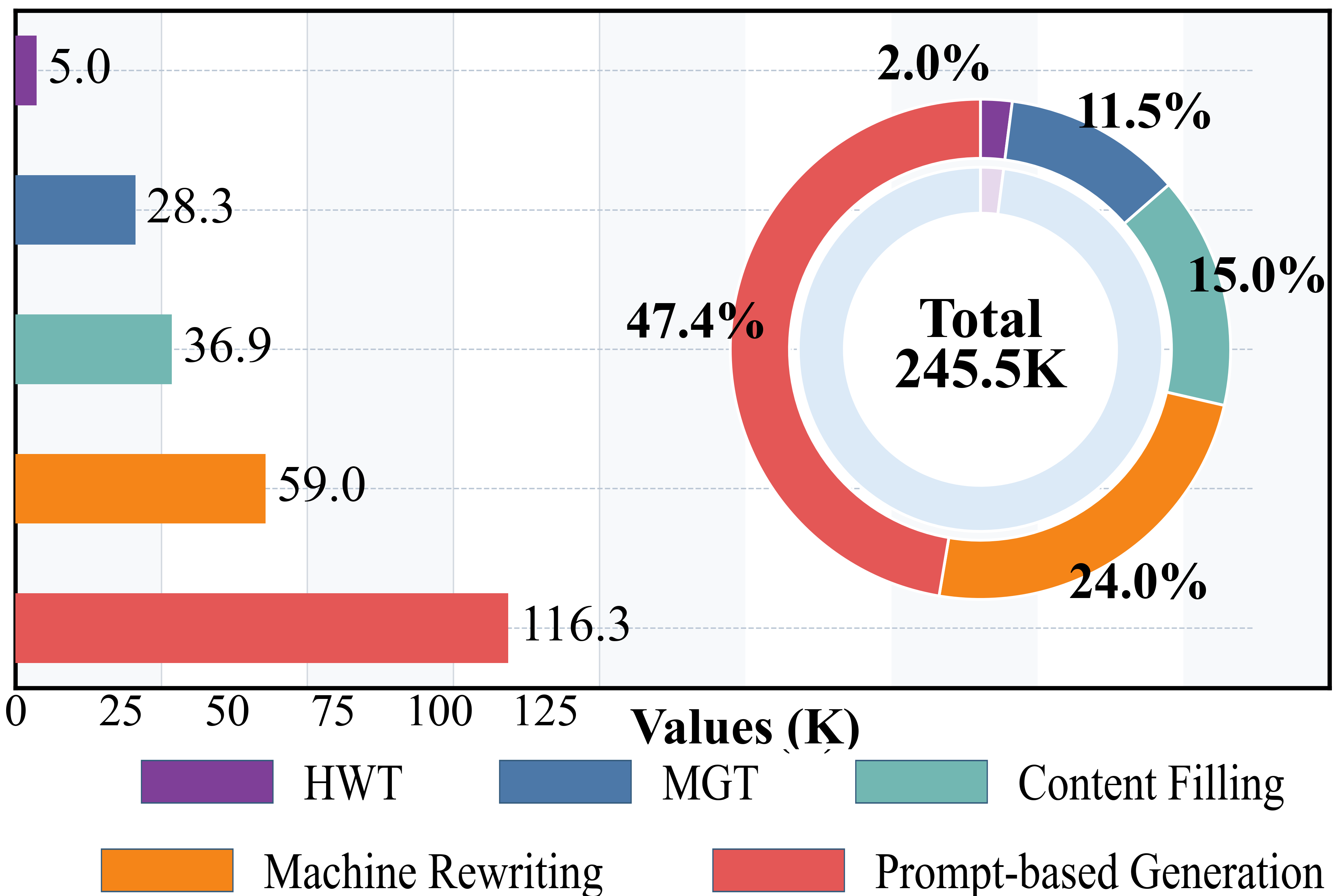}
    \caption{Composition of C-HAT-240K across production settings.}
    \label{fig:text-composition}
    \vspace{-15pt}
\end{wrapfigure}
then remove empty or malformed outputs, API error responses, prompt leakage, residual Markdown/HTML wrappers, severe encoding corruption, and clearly truncated or degenerated generations. 
The resulting corpus contains 245,539 records, including 5,000 shared HWT samples and 240,539 machine-involved samples spanning the benchmark's four production settings, as summarized in Fig.~\ref{fig:text-composition}. Finally, the retained records are mapped to a unified schema, with text fields and metadata such as source identifiers, domains, production types, subtype labels, and generators standardized and validated. Detailed dataset statistics are provided in Appendix~\ref{app:dataset-statistics}.

\subsection{Benchmark Setting}
\label{sec:Detection-Methods-and-Evaluation}

\noindent\textbf{Evaluation Tasks.}
To characterize detector behavior under different sources of variation, C-HAT-Bench evaluates four complementary tasks:

\begin{itemize}[
    leftmargin=1.2em,
    labelindent=0pt,
    labelwidth=0.7em,
    labelsep=0.5em,
    itemsep=2pt,
    topsep=2pt
]

\item \textbf{Detection performance across human--AI collaboration conditions.} We evaluate document-level detection across collaboration conditions to measure how the form and extent of machine involvement affect detectability.

\item \textbf{Detection sensitivity across text domains.} We compare detectors within each text-production setting to determine how detection performance depends on the source domain.

\item \textbf{Boundary localization in human--AI collaborative texts.} We identify transitions between human-written and machine-generated segments to assess whether detectors can recover fine-grained authorship changes.

\item \textbf{Cross-condition generalization.} We evaluate transfer across domains, collaboration modes, and generation models to measure detector robustness under distribution shifts.

\end{itemize}

\noindent\textbf{Detectors.}
To cover general and task-specific detection settings, we use 10 zero-shot and 7 off-the-shelf supervised detectors. The first two experiments evaluate the same detector sets. Boundary localization uses WCP and DeBERTa--BiGRU--CRF, while generalization experiments use AIGC-detector-zhv2 and IMBD. Detailed information on the detectors is provided in Appendix~\ref{app:detector-details}.

\noindent\textbf{Evaluation Metrics.}
To align each metric with the corresponding task, text-level detection is evaluated using AUROC~\citep{fawcett2006introduction}. Accuracy, and F1-score~\citep{sokolova2009systematic} at a fixed $5\%$ FPR operating point. Boundary localization uses sentence-level Cohen's $\kappa$~\citep{cohen1960coefficient}, WindowDiff~\citep{pevzner2002critique}, and Boundary F1@1~\citep{zeng2024sentences}. For generalization, we report accuracy at cell-specific thresholds maximizing Youden's $J$ on labeled test samples, a criterion also used for zero-shot calibration in TSM-Bench~\citep{quaremba2026tsmbench}.

\edef\chactSavedTopnumber{\number\value{topnumber}}
\setcounter{topnumber}{1}


\begin{table}[t]
\centering
\scriptsize
\setlength{\tabcolsep}{2.4pt}
\renewcommand{\arraystretch}{1.04}

\definecolor{MGT}{HTML}{F6E4E4}
\definecolor{contentfill}{HTML}{E7EEF8}
\definecolor{rewriting}{HTML}{E8F3E8}
\definecolor{promptgen}{HTML}{FFF1D6}
\definecolor{bestresult}{HTML}{FFD966}
\definecolor{secondresult}{HTML}{D9D9D9}

\newcommand{\best}[1]{\cellcolor{bestresult}#1}
\newcommand{\secondbest}[1]{\cellcolor{secondresult}#1}

\newcolumntype{P}{>{\columncolor{MGT}\centering\arraybackslash}X}
\newcolumntype{C}{>{\columncolor{contentfill}\centering\arraybackslash}X}
\newcolumntype{R}{>{\columncolor{rewriting}\centering\arraybackslash}X}
\newcolumntype{G}{>{\columncolor{promptgen}\centering\arraybackslash}X}

\caption{Detector performance under human--AI collaboration, measured by AUROC. The best two results are highlighted by \colorbox{bestresult}{\strut 1st} and \colorbox{secondresult}{\strut 2nd}. Here, MGT denotes the prefix-conditioned continuation condition; LC/HC indicate low/high coverage (15--35\%/50--75\%), and C/D indicate single-suffix/separated-span filling; Light/Deep denote expression polishing/structural rewriting; and L1--L3 indicate increasing prompt detail, with L3+Ref additionally using a same-domain reference.}
\label{tab:finegrained_auroc}
\begin{tabularx}{\linewidth}{@{}l*{1}{P}*{4}{C}*{2}{R}*{4}{G}@{}}
\toprule
\rowcolor{white}
\multirow{2}{*}{\textbf{Detector}}
& \multicolumn{1}{c}{\textbf{MGT}}
& \multicolumn{4}{c}{\textbf{Content Filling}}
& \multicolumn{2}{c}{\textbf{Machine Rewriting}}
& \multicolumn{4}{c}{\textbf{Prompt-based Generation}} \\
\cmidrule(lr){2-2}
\cmidrule(lr){3-6}
\cmidrule(lr){7-8}
\cmidrule(lr){9-12}

\rowcolor{white}
& {}
& LC-C
& LC-D
& HC-C
& HC-D
& Light
& Deep
& L1
& L2
& L3
& L3+Ref. \\
\midrule

\multicolumn{12}{@{}l@{}}{\textit{Zero-shot detectors}} \\
\midrule

Log-Likelihood
& 0.743 & 0.651 & 0.631 & 0.715 & 0.695
& 0.634 & 0.629 & 0.758 & 0.730 & 0.694 & 0.697 \\

LogRank
& 0.746 & 0.644 & 0.624 & 0.709 & 0.689
& 0.623 & 0.616 & 0.762 & 0.727 & 0.688 & 0.691 \\

Entropy
& 0.317 & 0.374 & 0.386 & 0.333 & 0.334
& 0.393 & 0.396 & 0.303 & 0.324 & 0.341 & 0.338 \\

DetectLLM-LRR
& 0.712 & 0.578 & 0.560 & 0.635 & 0.610
& 0.529 & 0.516 & 0.723 & 0.661 & 0.606 & 0.608 \\

Fast-DetectGPT
& 0.725 & 0.637 & 0.618 & 0.671 & 0.635
& 0.585 & 0.525 & 0.727 & 0.608 & 0.600 & 0.606 \\

Binoculars
& 0.874 & 0.634 & 0.579 & 0.770 & 0.673
& 0.626 & 0.648 & 0.867 & 0.794 & 0.752 & 0.756 \\

LastDE
& 0.835 & 0.681 & 0.664 & 0.743 & 0.747
& 0.671 & 0.715 & 0.848 & 0.829 & 0.773 & 0.769 \\

LastDE++
& 0.859 & 0.707 & 0.669 & 0.798 & 0.749
& 0.697 & 0.683 & 0.857 & 0.801 & 0.752 & 0.755 \\

SpecDetect
& \best{0.895}
& \best{0.777}
& \best{0.748}
& \best{0.851}
& \best{0.829}
& \best{0.763}
& \best{0.773}
& \best{0.906}
& \best{0.882}
& \best{0.842}
& \best{0.843} \\

SpecDetect++
& \secondbest{0.887}
& \secondbest{0.749}
& \secondbest{0.714}
& \secondbest{0.828}
& \secondbest{0.791}
& \secondbest{0.748}
& \secondbest{0.749}
& \secondbest{0.894}
& \secondbest{0.851}
& \secondbest{0.806}
& \secondbest{0.807} \\

\midrule
\multicolumn{12}{@{}l@{}}{\textit{Supervised detectors}} \\
\midrule

RADAR
& 0.540 & 0.373 & 0.405 & 0.311 & 0.300
& 0.434 & 0.464 & 0.583 & 0.351 & 0.370 & 0.378 \\

ReMoDetect
& \secondbest{0.833}
& \secondbest{0.706}
& \secondbest{0.679}
& \secondbest{0.798}
& \secondbest{0.761}
& \best{0.730}
& \best{0.800}
& \secondbest{0.879}
& \best{0.855}
& \best{0.807}
& \best{0.804} \\

Longformer
& 0.826 & 0.546 & 0.514 & 0.578 & 0.531
& 0.522 & 0.542 & 0.877 & 0.683 & 0.649 & 0.644 \\

BLOOMZ-3B
& \best{0.979}
& \best{0.808}
& \best{0.746}
& \best{0.817}
& \best{0.768}
& \secondbest{0.668}
& 0.634
& \best{0.998}
& \secondbest{0.786}
& \secondbest{0.771}
& \secondbest{0.770} \\

OpenAI RoBERTa
& 0.480 & 0.421 & 0.414 & 0.424 & 0.350
& 0.541 & 0.582 & 0.482 & 0.483 & 0.487 & 0.494 \\

ConDA
& 0.501 & 0.503 & 0.487 & 0.526 & 0.468
& 0.471 & 0.479 & 0.557 & 0.638 & 0.587 & 0.581 \\

ChatGPT Detector
& 0.724 & 0.599 & 0.545 & 0.724 & 0.640
& 0.652 & \secondbest{0.694}
& 0.741 & 0.736 & 0.686 & 0.693 \\

\bottomrule
\end{tabularx}
\vspace{-10pt}
\end{table}

\section{EXPERIMENTAL RESULTS \& DISCUSSION}
\label{sec:experiments}

Our evaluation uses four questions: \textbf{RQ1 (Detection Performance across Human--AI Collaboration Conditions):} How does detector performance change with the form and degree of human--AI collaboration? (Sec.~\ref{sec:fine_grained}) \textbf{RQ2 (Domain sensitivity):} How do text domains affect the detection of human--AI collaborative text? (Sec.~\ref{sec:detection_sensitivity}) \textbf{RQ3 (Boundary localization):} Can existing detectors accurately localize the boundaries between human-written and machine-generated content in collaborative text? (Sec.~\ref{sec:boundary_localization}) \textbf{RQ4 (Cross-distribution generalization):} Can supervised detectors generalize across shifts in domain, collaboration type, and generation model? (Sec.~\ref{sec:generalization}) Detailed experimental settings are provided in Appendix~\ref{app:experimenta-setup}.

\begin{figure}[!b]
    \centering
    \includegraphics[width=0.95\textwidth]{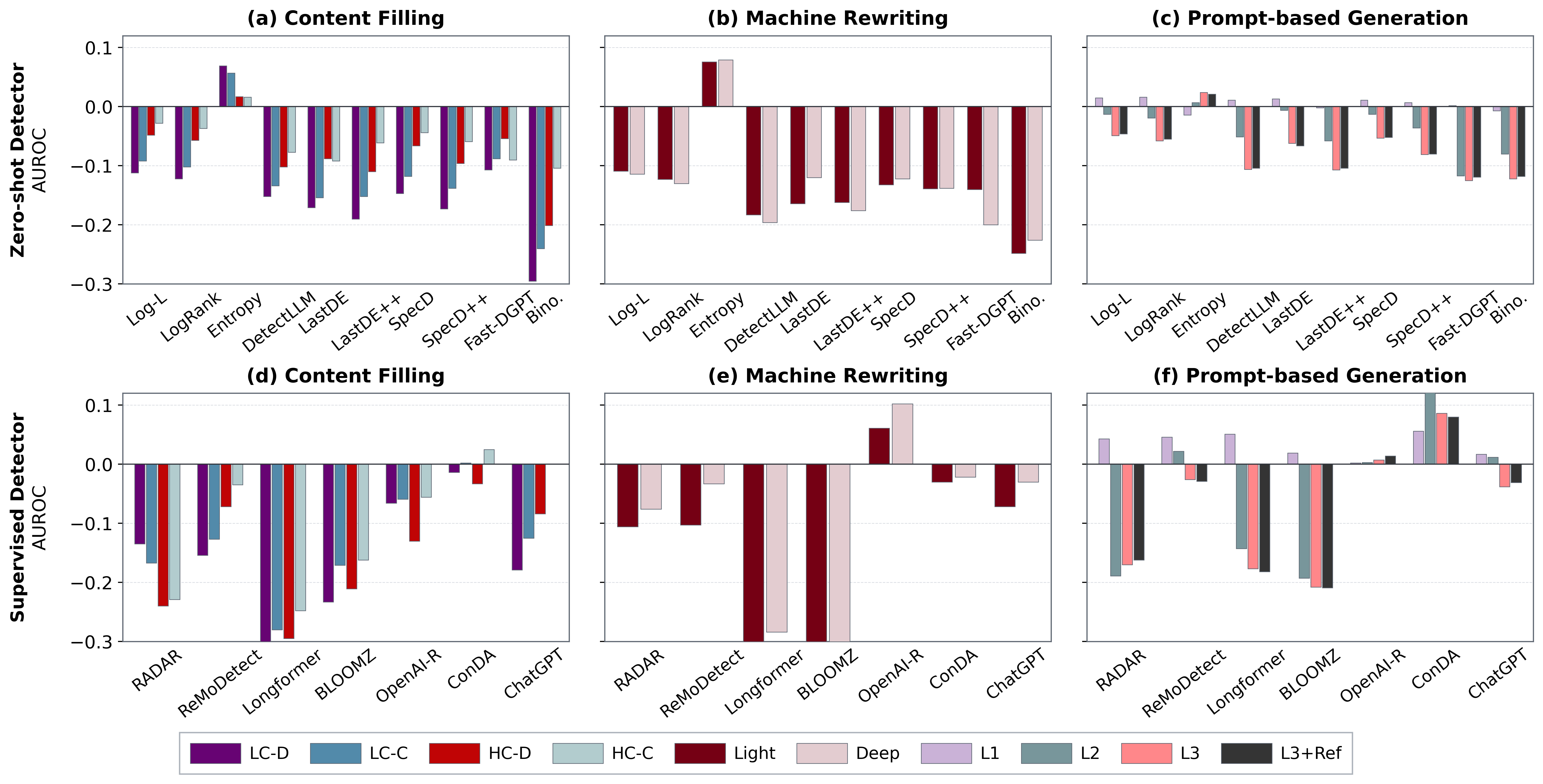}
    \caption{Detector performance relative to MGT across human--AI collaboration conditions. Top and bottom rows show zero-shot and supervised detectors, respectively.}
    \label{fig:finegrained_collaboration}
    \vspace{-10pt}
\end{figure}

\subsection{Detection Performance across Human--AI Collaboration Conditions (RQ1)}
\label{sec:fine_grained}

\noindent\textbf{Experiment design.}
We evaluate 10 zero-shot and 7 off-the-shelf supervised detectors on C-HAT-Bench for document-level binary classification.
To avoid truncation under detector input-length limits, we use 512-token sliding windows and average window-level scores.
We conduct a stride sensitivity analysis and adopt strides of 256 and 410 tokens for zero-shot and supervised detectors, respectively. Further details of the sliding-window hyperparameter analysis are provided in Appendix~\ref{app:sliding-window}. For each detector and condition, we report the mean \textsc{AUROC}, Aggregate results weight detectors and the ten collaboration conditions equally, with relative decreases measured against the corresponding mean MGT AUROC. with MGT as the reference condition. Additional experimental results are reported in Appendix~\ref{app:fine-grained-results}.

\noindent\textbf{Experimental results.}
Table~\ref{tab:finegrained_auroc} reports the mean AUROC of zero-shot and supervised detectors for MGT and each fine-grained collaboration condition. Figure~\ref{fig:finegrained_collaboration} visualizes the corresponding AUROC differences relative to MGT across the three collaboration subsets.

\noindent\textbf{\underline{Observation~\ding{182}:} Human--AI collaboration changes the detectability of machine involvement text.}
Relative to MGT, mean AUROC decreases by 11.1\% and 13.3\% for zero-shot and supervised detectors, respectively, when human--AI collaboration conditions are equally weighted (Table~\ref{tab:finegrained_auroc}). Content Filling and Machine Rewriting reduce mean AUROC in both groups, while Prompt-based Generation shifts from performance close to MGT at L1 to lower performance at L3. Content mixing may dilute machine-generation cues at the document level, while rewriting and richer prompts may alter the lexical and stylistic patterns used for detection. The comparable results for L3 and L3+Ref. suggest that reference text has limited additional impact under already detailed prompt constraints. Thus, MGT performance alone cannot capture detection difficulty across collaboration conditions, which also depends on the form and extent of human intervention.

\noindent\textbf{\underline{Observation~\ding{183}:} Zero-shot detectors exhibit consistent responses to human--AI collaboration conditions.}
Within each collaboration subset, zero-shot detectors generally exhibit similar performance trends across human--AI collaboration conditions (Figure~\ref{fig:finegrained_collaboration}(a)--(c)). We attribute this consistency to their shared reliance on language-model statistics, including token probabilities, token rankings, and sequence-level distributional features. Collaboration may reduce the discriminative value of these shared cues, affecting multiple detectors despite differences in their scoring functions. Thus, diversity in scoring functions does not necessarily imply diversity in detection evidence; shared statistical dependencies can create common vulnerabilities to collaborative text.

\noindent\textbf{\underline{Observation~\ding{184}:} Supervised detectors exhibit method-dependent responses to human--AI collaboration conditions.}
As shown in Figure~\ref{fig:finegrained_collaboration}(d)--(f), supervised detectors exhibit different, sometimes opposing, performance changes under the same collaboration conditions. We attribute this variation to differences in training corpora, architectures, and learning objectives, which shape detector-specific decision boundaries. A collaboration-induced feature change may therefore weaken one detector's learned cues while preserving or reinforcing those used by another. The absence of a consistent difficulty ordering across the evaluated detectors suggests that detectability depends on the interaction between collaboration conditions and learned decision criteria.

\begin{figure}[t]
    \centering
    \includegraphics[width=0.95\textwidth]{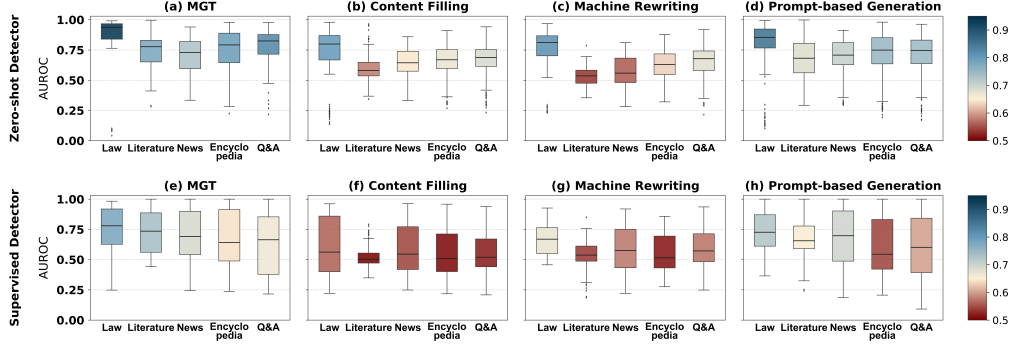}
    \caption{AUROC distributions across five text domains for zero-shot (top) and supervised (bottom) detectors on MGT and human--AI collaborative text. Box colors indicate median AUROC.}
    \label{fig:domain_auroc_boxplot}
    \vspace{-10pt}
\end{figure}

\subsection{Detection Sensitivity Across Text Domains (RQ2)}
\label{sec:detection_sensitivity}

\noindent\textbf{Experiment design.}
To assess the impact of text domains on detection performance, we compare zero-shot and supervised detectors across C-HAT-240k's five domains under MGT and three human--AI collaboration settings. Additional experimental results are provided in Appendix~\ref{app:domain_results}.

\noindent\textbf{Experimental results.}
Figure~\ref{fig:domain_auroc_boxplot} presents the AUROC distributions of the two detector families across different text domains. Subfigures~(a)--(d) correspond to zero-shot detectors, while Subfigures~(e)--(h) correspond to supervised detectors. We derive the following key observations.

\noindent\textbf{\underline{Observation~\ding{185}:} Text domains exhibit stable but uneven detection difficulty.}
As shown in Figure~\ref{fig:domain_auroc_boxplot}, the law domain achieves the highest median AUROC in all eight subfigures, whereas literature and news generally yield lower detection performance, with this pattern remaining broadly stable across production settings. This consistency suggests that domain characteristics systematically shape the separability of human and machine text: the constrained terminology and discourse conventions of legal writing may expose generation artifacts, whereas the greater expressive variation in open-ended domains can increase feature overlap. Thus, text domain constitutes a systematic source of variation in detection difficulty rather than a secondary evaluation factor.

\noindent\textbf{\underline{Observation~\ding{186}:} Zero-shot and supervised detectors exhibit distinct domain sensitivities.}
Zero-shot detectors preserve a stable domain ordering across collaboration settings, with an average IQR of 0.18, whereas supervised detectors show more dispersed and less consistent distributions, with the average IQR increasing to 0.31 and the weakest domain varying across settings. This difference reflects their underlying evidence: zero-shot detectors rely on shared probability and token-ranking signals, while supervised detectors learn decision boundaries shaped by their training corpora and architectures. Consequently, aggregate scores can conceal domain-specific weaknesses, making cross-domain coverage essential for assessing detector reliability.

\subsection{Boundary Localization in Human--AI Collaborative Texts (RQ3)}
\label{sec:boundary_localization}

\noindent\textbf{Experiment design.} We evaluate the fine-tuned WCP and the DeBERTa–BiGRU–CRF on the Content 
Filling subset. By aligning masking plans with the final texts, we recover machine-generated 
\begin{wrapfigure}{r}{0.60\textwidth}
    \vspace{-10pt}
    \centering
    \includegraphics[width=\linewidth]{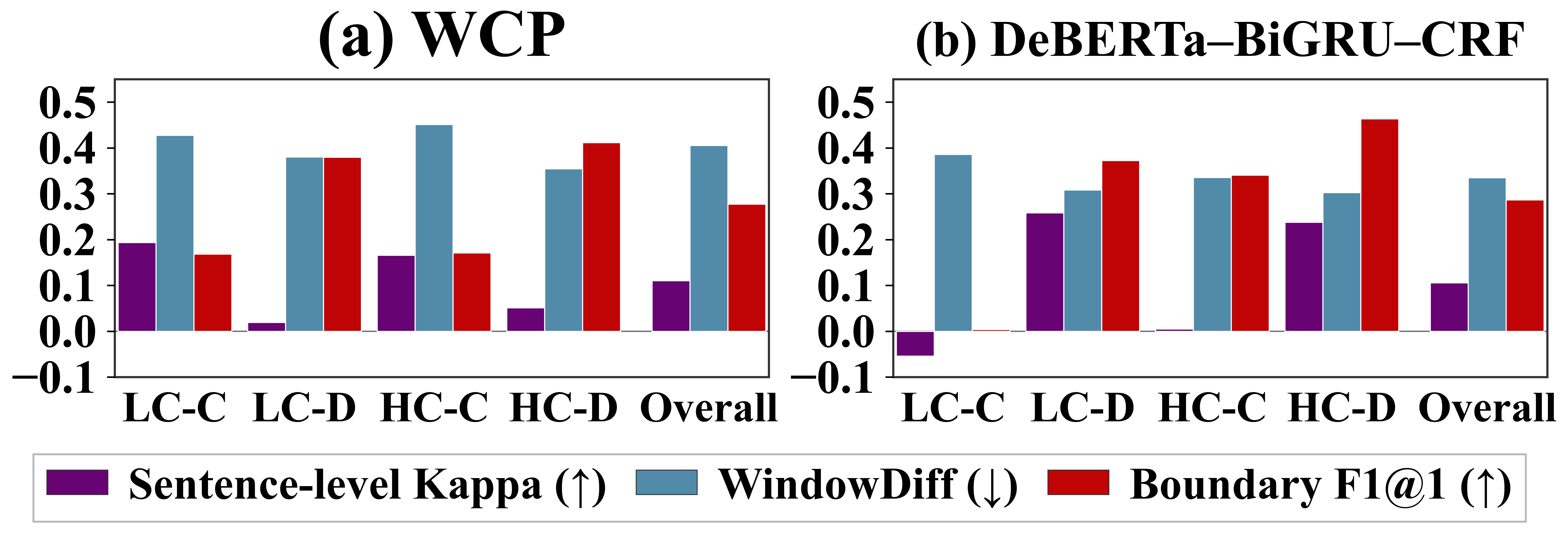}
    \caption{Boundary-localization performance of WCP and DeBERTa–BiGRU–CRF across Content Filling conditions.}
    \label{fig:bouandry-detect}
    \vspace{-10pt}
\end{wrapfigure}
spans and derive sentence-level authorship labels and gold boundaries. We report sentence-level $\kappa$, WindowDiff, and Boundary F1@1, while additional preprocessing details and experimental results are provided in the appendix~\ref{app:boundary-localization}.

\noindent\textbf{Experimental results.}
Figure~\ref{fig:bouandry-detect} reports sentence-level $\kappa$, WindowDiff, and Boundary F1@1 for WCP and the DeBERTa–BiGRU–CRF in panels (a) and (b), respectively.

\noindent\underline{\textbf{Observation~\ding{187}:}}
\textbf{Boundary matching and authorship recovery respond differently to boundary density.}
Across coverage levels, WCP achieves higher Boundary F1@1 under dispersed filling but lower sentence-level $\kappa$ than under contiguous filling. WCP proposes roughly four change points per text, whereas contiguous texts have one gold boundary and dispersed texts have about five. This mismatch makes contiguous texts prone to oversegmentation and partly explains higher boundary-matching scores under dispersed filling. Because WCP assigns sentence labels separately from change-point proposals, matching a boundary does not ensure authorship assignment on either side.

\noindent\underline{\textbf{Observation~\ding{188}:}}
\textbf{Low-coverage contiguous filling exposes opposing failure modes.}
DeBERTa–BiGRU–CRF's Boundary F1@1 falls to 0.004, compared with 0.341 at high coverage; 70.9\% of these samples yield no predicted boundary. As its token predictions are mapped to sentence-level labels, this pattern points to missed authorship transitions rather than small boundary offsets. WCP, in contrast, tends to predict excess boundaries on the same texts.

\subsection{Cross-Domain Generalization of AIGC-detector-zhv2 (RQ4)}
\label{sec:generalization}

\def\CDWidth{0.50\textwidth} 
\def\CDScale{1.00}          
\def\CDBottomAdjust{-14pt}   

\noindent\textbf{Experiment design.}
We train AIGC-detector-zhv2 separately on each of five domains and evaluate it across all five. Diagonal and off-diagonal cells denote in-domain and cross-domain tests. Each cell reports ACC using a Youden's $J$ threshold selected from its labeled test samples; the results therefore reflect target-calibrated performance. Further details appear in Appendix~\ref{app:cross-detection}.

\medskip

\begin{wrapfigure}{r}{0.46\textwidth}
    \vspace{-10pt}
    \centering
    \includegraphics[width=\linewidth]{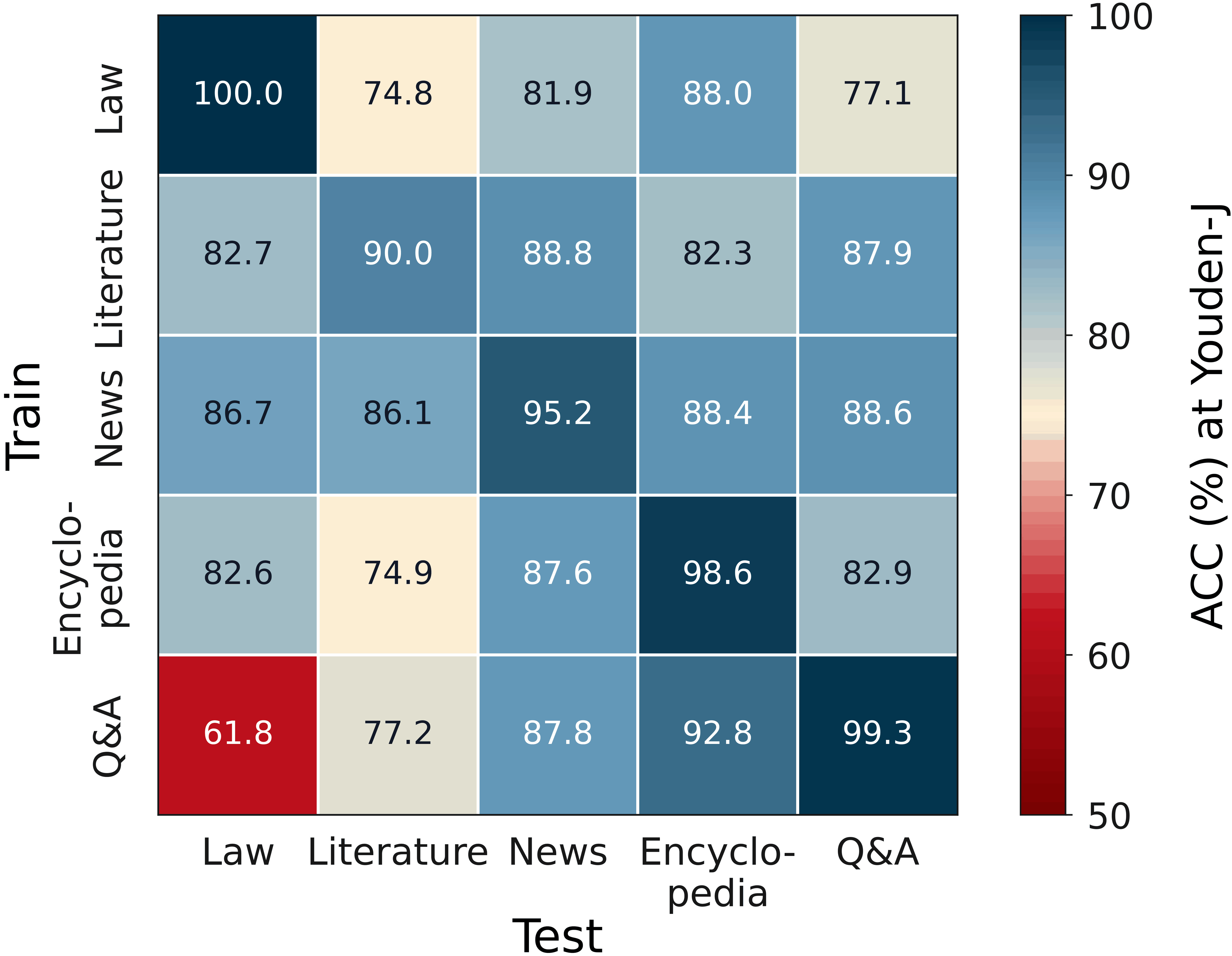}
    \caption{Cross-domain accuracies of AIGC-detector-zhv2 across five text domains.}
    \label{fig:cross_domain}
    \vspace{-20pt}
\end{wrapfigure}

\noindent\textbf{Experimental results.}
Figure~\ref{fig:cross_domain} presents the resulting $5\times5$ ACC matrix for AIGC-detector-zhv2.
Rows and columns denote the training and test domains, respectively; each row reflects source-domain transferability, while each column reflects target-domain difficulty. 

\medskip

\noindent\underline{\textbf{Observation~\ding{189}:}}
\textbf{News-trained AIGC-detector-zhv2 exhibits the strongest cross-domain transferability.}
Training on News produces the highest mean cross-domain ACC (87.45\%), with 86.1\%--88.6\% on the other domains, whereas Q\&A yields 79.9\% and falls to 61.8\% on Law.
This contrast indicates that transferability is shaped by the specificity of the learned cues, rather than by the magnitude of the domain shift alone.
News may encourage more domain-agnostic cues through its 
broader lexical and discourse variation, whereas Q\&A may induce patterns tied to question--answer. 
Thus, the source domain should be treated as an explicit factor in cross-domain evaluation, since 
these structures that transfer poorly to formal domains such as Law.
High in-domain accuracy alone does not establish transferable detection cues.

\medskip

\noindent\underline{\textbf{Observation~\ding{190}:}}
\textbf{Literature is a target-domain bottleneck but a relatively transferable source domain.}
For AIGC-detector-zhv2, the mean cross-domain ACC is lowest when Literature is the target domain (78.24\%), but reaches 85.43\% when the detector is trained on Literature and evaluated on the remaining domains.
This asymmetry shows that Literature is not uniformly incompatible with the other domains: incoming transfer is difficult, whereas outgoing transfer remains relatively strong.
Its greater lexical and stylistic variation may obscure cues learned from other domains, while training on such variation may encourage a broader decision rule.
Thus, cross-domain generalization is directional, and aggregate performance can conceal target-specific bottlenecks.

\section{Conclusion and Future Directions}

We present C-HAT-Bench, a benchmark for studying Chinese human--AI collaborative text detection. By linking shared human-written sources to machine-involved variants under controlled production settings, the benchmark enables detection behavior to be examined as collaboration granularity, source domain, and generative model vary. Our experiments show that detector performance is substantially affected by collaborative conditions and domain shifts, revealing limitations that are difficult to observe in evaluations based only on fully human-written and fully machine-generated text. We hope C-HAT-Bench will encourage the field to treat human--AI collaboration as a standard evaluation setting and promote the development of detection methods that remain reliable as writing processes and data distributions change.

These findings motivate three directions for future research:

\begin{itemize}[
    label=\textbullet,
    leftmargin=0.9em,
    labelwidth=0.45em,
    labelsep=0.35em,
    itemindent=0pt,
    align=left,
    itemsep=3pt,
    topsep=3pt,
    parsep=0pt,
    partopsep=0pt
]

\item \textbf{Detection under human--AI collaboration.}
A document that retains human-written context and contains only partial model intervention may not exhibit the same cues as fully machine-generated text. As the form and extent of collaboration change, the boundary between human and machine authorship may become less explicit, making collaborative text detection a fundamental unresolved problem.

\item \textbf{Generalization beyond the evaluation condition.}
Our generalization experiments show that detector performance deteriorates substantially on out-of-domain data, indicating that strong results under the original distribution do not guarantee reliable detection under other conditions. Improving generalization performance should therefore remain an important direction for future research, so that detector decisions remain dependable when the evaluation data differ from those used during development.

\item \textbf{Sensitivity to collaboration versus false positives.}
It is also possible that greater sensitivity to subtle machine involvement may increase false positives on high-quality human-written text. Such texts could contain regular or highly polished linguistic patterns that resemble machine-produced text even when no model has participated. Whether this trade-off exists, which types of human writing are most vulnerable, and how it varies across collaboration conditions remain open questions for future investigation.

\end{itemize}

\noindent\textbf{Limitations.}\quad
The current benchmark focuses on controlled fine-grained collaboration settings and does not yet examine more complex collaboration patterns or incorporate feedback loops into text data construction. We hope future research will extend this work toward richer collaboration processes and feedback-aware benchmark design.

\section*{AI Use Statement}

During manuscript preparation, generative AI tools were used only for limited language polishing and presentation-related assistance. They were not used to make scientific decisions, fabricate experimental results, or create citations. The authors reviewed the manuscript and take responsibility for the study design, analyses, claims, and final text.

\section*{Ethics Statement}

This work uses existing Chinese text corpora and model-generated text variants and does not involve human-subject experiments. Source materials and released models are used subject to their applicable licenses and terms. The benchmark is intended for research and evaluation. Because detection performance can vary across domains, writing styles, and generation conditions, and because false positives may affect human-written text, detector outputs should be treated as supporting evidence and interpreted with human judgment rather than used as the sole basis for disciplinary or other consequential decisions.

\bibliography{iclr2027_conference}
\bibliographystyle{iclr2027_conference}

\clearpage
\appendix

\clearpage
\appendix
\startcontents[appendix]

\begingroup
\normalfont\normalsize
\hypersetup{hidelinks}

\section*{Appendix}
\subsection*{Table of Contents}

\titlecontents{section}[2em]
  {\addvspace{0.6em}\normalfont\normalsize\scshape}
  {\contentslabel{1.5em}}
  {}
  {\hfill\normalfont\normalsize\contentspage}

\titlecontents{subsection}[4em]
  {\normalfont\normalsize}
  {\contentslabel{2.5em}}
  {}
  {\titlerule*[0.7em]{.}\contentspage}

\printcontents[appendix]{}{1}{\setcounter{tocdepth}{2}}

\endgroup
\clearpage

\section{Dataset Construction}
\label{app:dataset-construction}

\subsection{Human-Written Text Collection}
\label{app:hwt-collection}

To ensure consistency in corpus size and basic text-length distributions across domains, we select 1{,}000 human-written texts from each domain. Specifically, we adopt a length-centered sampling strategy: using approximately 1{,}250 characters as the target length, we first prioritize candidate texts close to this length and then gradually expand toward shorter and longer texts until 1{,}000 samples are obtained for each domain. All selected texts are further required to satisfy the Chinese-character proportion criterion, yielding 5{,}000 HWT samples in total.

\textbf{News}\quad THUCNews is a Chinese news classification corpus compiled from historical articles collected through Sina News RSS feeds between 2005 and 2011. It contains approximately 740{,}000 articles spanning 14 topical categories. For C-HAT-240k, we select news texts from six categories: finance, education, technology, society, politics, and sports. To ensure that Chinese is the predominant language, we further exclude articles containing an excessive proportion of English-language content.

\textbf{Law}\quad LeCaRDv2 is a large-scale Chinese legal case retrieval dataset comprising 800 query cases and 55{,}192 candidate cases drawn from approximately 4.3 million criminal case documents, with broad coverage of criminal charges \citep{li2024lecardv2}. We use its candidate-case collection and extract the \texttt{qw} field containing the full text of each case document, rather than concatenating the separately provided fields for case facts, judicial reasoning, and judgment outcomes. Case numbers, dates, statutory provisions, and other domain-specific information are retained.

\textbf{Q\&A}\quad Zhihu-KOL is a Chinese question-answering corpus collected from Zhihu, comprising 1{,}006{,}218 question--answer pairs whose responses were authored by key opinion leader (KOL) users. We use a subset of 50{,}000 pairs as the candidate pool. Because the original corpus contains a substantial number of recently posted responses, we retain only the 25{,}833 pairs posted before 2020, as determined by the \textit{answer\_creation\_time} field, to reduce the risk of machine-generated content entering the HWT corpus. Responses containing an excessive proportion of English-language content are further excluded. For HWT construction, the response in the \textit{RESPONSE} field serves as the evaluation text, while the corresponding question in the \textit{INSTRUCTION} field is retained to preserve the question--answer association and support data traceability; the question itself is not included in the detector input.

\textbf{Encyclopedia}\quad The encyclopedia texts are drawn from the Chinese Wikipedia dataset on ModelScope, which is compiled from Chinese Wikipedia entries and contains approximately 2.87 million open-domain encyclopedia articles. The corpus covers a wide range of entry types, including people, events, organizations, locations, concepts, and knowledge-oriented topics. Compared with news, legal, and question-answering texts, Wikipedia articles place greater emphasis on objective exposition, conceptual explanation, and factual description, making them suitable for constructing the encyclopedia-domain HWT subset. We select entries with sufficiently complete body text that are primarily written in Chinese and satisfy the length requirement, and further remove samples containing an excessive proportion of English-language content.

\textbf{Literature}\quad The literature texts are drawn from a locally curated collection of modern Chinese literary works. We select 102 modern literary works covering multiple genres, including novels, essays, prose writings, letters, and drama. Candidate passages are segmented according to natural paragraph boundaries and chapter structure, after which tables of contents, copyright information, empty paragraphs, and passages that do not satisfy the length requirement are removed. To prevent a small number of long works from dominating the subset, we control the number of samples drawn from each work and finally select 1{,}000 continuous passages from different works as the literature-domain HWT subset.

\subsection{Data Cleaning and Validation}
\label{app:data-cleaning}

\begin{table}[H]
\centering
\footnotesize
\setlength{\tabcolsep}{4pt}
\renewcommand{\arraystretch}{1.15}

\caption{HWT cleaning and validation rules for C-HAT-240K.}
\label{tab:app-cleaning-validation}

\begin{tabularx}{\linewidth}{
    @{}
    >{\raggedright\arraybackslash}p{0.16\linewidth}
    >{\raggedright\arraybackslash}p{0.18\linewidth}
    >{\raggedright\arraybackslash}X
    @{}
}
\toprule
\textbf{Stage} & \textbf{Criterion} & \textbf{Rule} \\
\midrule

Human-text selection
& Length-based selection
& Starting from 1250 characters, progressively expand the selection range toward shorter and longer texts until 1,000 samples are collected per domain. \\
\addlinespace[3pt]

Human-text selection
& Chinese-character ratio
& For news, law, Q\&A, and encyclopedia, start at 90\% and relax the threshold when insufficient samples are available, with a lower bound of 60\%. The denominator excludes whitespace. \\
\addlinespace[3pt]

Format cleaning
& Whitespace and markup
& Normalize line endings, remove redundant whitespace and invisible characters, and strip HTML/Markdown markup, code fences, and recognizable task-description prefixes. Text lengths are computed after cleaning. \\
\addlinespace[3pt]

\bottomrule
\end{tabularx}
\end{table}

\FloatBarrier

\begin{figure}[!htbp]
    \centering
    \includegraphics[width=\linewidth]
    {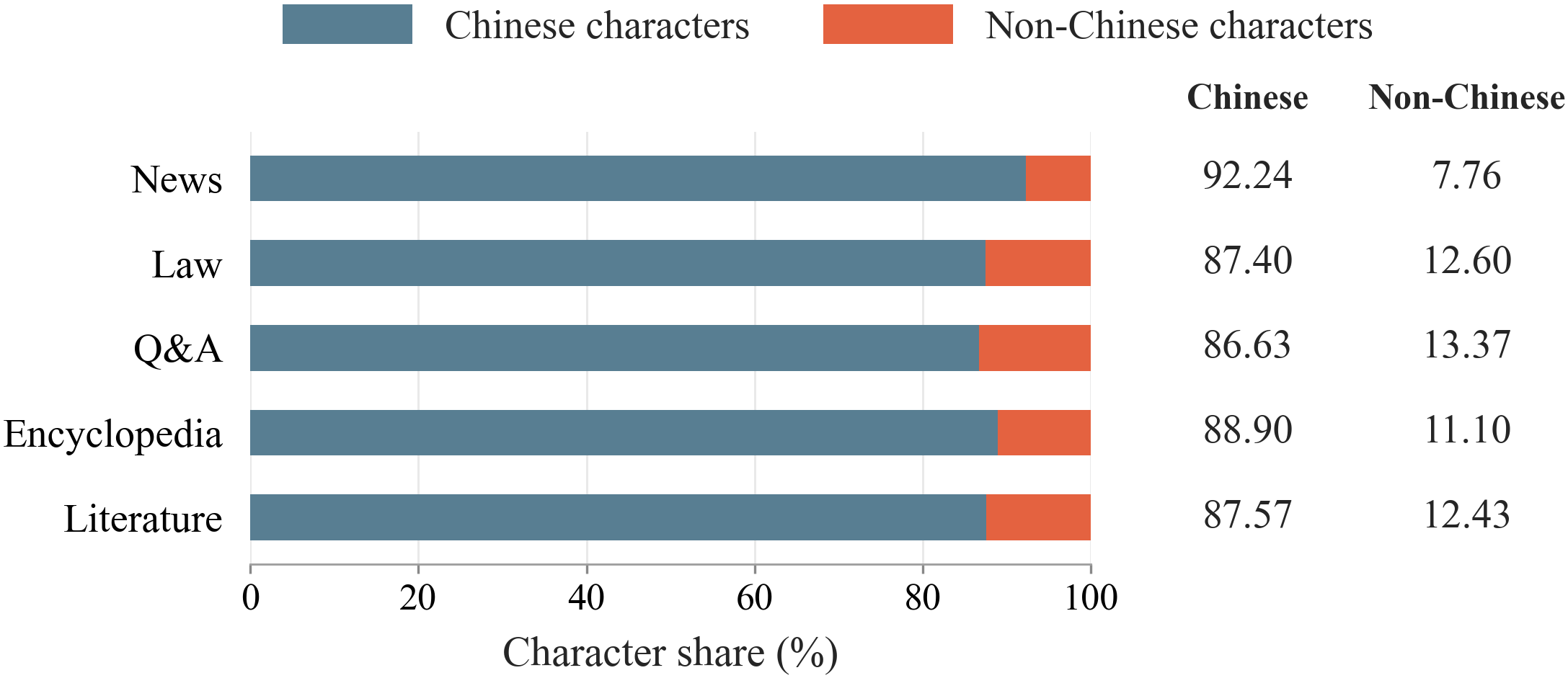}
    \caption{Character composition of the selected human corpus.
    Percentages are averaged over 1,000 texts per domain,
    excluding whitespace.}
    \label{fig:human-character-composition}
\end{figure}

\FloatBarrier

Table~\ref{tab:app-cleaning-validation} summarizes the rules for
human-text selection, generated-text cleaning, and record validation.
Figure~\ref{fig:human-character-composition} shows that the selected
human corpus is predominantly Chinese, with mean Chinese-character
proportions ranging from 86.63\% to 92.24\% across the five domains.
Non-Chinese characters include English letters, digits, and punctuation.

\subsection{Generative Models}
\label{app:generative-models}

We use the following generative models to construct MGT and human–AI collaborative samples. The model names refer to the exact
versions used in our experiments. Family-level technical reports are
cited where available, while official documentation is provided in
footnotes for the remaining models.

\textbf{GPT-5.5}\quad
An OpenAI GPT-series model used to generate machine-involved samples in our benchmark.\footnote{OpenAI, \textit{Models}, official model documentation. Available at: \url{https://platform.openai.com/docs/models} (accessed September 25, 2026).}

\textbf{Qwen3.7-Plus}\quad
A model in Alibaba's Qwen series used for Chinese and multilingual text
generation, with its family-level technical background described in the
Qwen3 technical report~\citep{yang2025qwen3}.

\textbf{GLM-5.2}\quad
A model in Z.ai's GLM series used for Chinese and English text
generation, with its family-level technical background described in the
GLM-5 technical report~\citep{glm5team2026glm5}.

\textbf{Kimi K2.6}\quad Model provided by Moonshot AI; the exact version is identified using the provider's official documentation.
\footnote{Moonshot AI, \textit{Kimi API Documentation}, official platform documentation.
Available at: \url{https://platform.moonshot.cn/docs/intro}
(accessed September 25, 2026).}

\textbf{DeepSeek V4 Pro}\quad
A model in the DeepSeek series used for text generation, with its
technical background described in the DeepSeek-V4 technical report
~\citep{deepseekai2026deepseekv4}.

\textbf{Gemini 3.5 Flash}\quad Gemini-series model provided by Google; the exact version is identified using the provider's official documentation.
\footnote{Google, \textit{Gemini API Documentation: Models}, official developer documentation.
Available at: \url{https://ai.google.dev/gemini-api/docs/models}
(accessed September 25, 2026).}

\subsection{Prompt Templates}
\label{app:prompt-templates}

\begin{CJK*}{UTF8}{gbsn}
\begin{promptbox}{MGT}

请根据下面这段文本开头进行续写。

要求：最终全文由“文本开头 + 续写内容”组成，全文长度应尽量接近原文长度。

原文长度：\texttt{\{human\_text\_length\}}

文本开头：\texttt{\{prefix\_text\}}

\end{promptbox}
\begin{promptbox}{Content filling}

\textbf{Continuous Span}

请根据 \texttt{masked\_text} 中 \texttt{[MASK\_1]} 前面的原文，自然续写 \texttt{[MASK\_1]} 对应的内容，使生成后的文本与前文保持自然连贯。

生成要求：

1）只生成 \texttt{[MASK\_1]} 对应内容，不要改写 \texttt{masked\_text} 中已经存在的原文。

2）不要复述或重复 \texttt{masked\_text} 中已经存在的连续内容。

3）续写内容应贴合原文的主题、事实、语气、文体和叙述逻辑，并自然承接前文。

4）\texttt{[MASK\_1]} 的 \texttt{generated\_text} 长度控制在 \texttt{min\_generate\_length} 到 \texttt{max\_generate\_length} 之间。

\medskip
\hrule
\medskip

\textbf{Dispersed Spans}

请根据 \texttt{current\_text} 中当前 MASK 所在位置的上下文，自然生成该 MASK 对应的内容，使填补后的文本在上下文之间保持自然连贯。

生成要求：

1）只生成 \texttt{current\_mask\_id} 对应内容，不要生成其他 MASK，也不要改写 \texttt{current\_text} 中已经存在的原文。

2）不要复述或重复 \texttt{current\_text} 中已经存在的连续内容。

3）生成内容应贴合原文的主题、事实、语气、文体和上下文逻辑，并自然连接 \texttt{current\_mask\_id} 所在位置的前后文。

4）\texttt{generated\_text} 长度控制在 \texttt{min\_generate\_length} 到 \texttt{max\_generate\_length} 之间。

5）如果 \texttt{current\_text} 中已有前序 MASK 的生成内容，请把这些已生成内容视为当前上下文的一部分，并在此基础上继续保持文本连贯。

\end{promptbox}
\begin{promptbox}{Machine Rewriting}

\textbf{Light Revision}

请对下面文本进行轻度表达润色。仅在局部层面优化用词、语序、衔接和句子流畅度，减少重复、生硬或不自然的表达。不要改变原文事实信息、核心含义、段落结构和整体写作风格，不要加入新的内容。只输出修改后的正文。

\medskip
\hrule
\medskip

\textbf{Deep Rewriting}

请对下面文本进行深度改写。在保留原文事实信息、主题内容和领域文体特征的基础上，可以重新组织句式结构、表达方式和段落内部逻辑，使文本整体更加自然、清晰、连贯且具有重新写作后的表达形式。不要添加原文未包含的新事实。只输出改写后的正文。

\end{promptbox}
\begin{promptbox}{Prompt-based Generation: News Templates}

\textbf{Level 1}

请根据下面的新闻主题写一篇中文新闻报道，文本长度应尽量接近指定长度。

长度要求：\texttt{\{target\_length\}}

新闻主题：\texttt{\{news\_title\}}

\medskip
\hrule
\medskip

\textbf{Level 2}

请写一篇新闻报道，主题是“\texttt{\{main\_subject\}}”。文章需要围绕主要事件展开：\texttt{\{main\_event\}}。事件结果或当前进展为：\texttt{\{result\_or\_status\}}。文本长度尽量接近\texttt{\{target\_length\}}。

\medskip
\hrule
\medskip

\textbf{Level 3}

你是一名擅长中文新闻报道写作的专业写作者。请根据以下任务大纲生成一篇新闻报道。正文应信息清楚、叙述完整、语言客观。只输出正文，文本长度尽量接近\texttt{\{target\_length\}}。

任务大纲：

1）写作思路：全文应主要参考新闻报道的叙事顺序和信息重心，按照“导语概括核心事实—交代时间地点与相关主体—展开事件经过—说明结果进展—补充背景或来源信息”的方式组织内容。

2）新闻导语：围绕新闻主题“\texttt{\{main\_subject\}}”写出开头，结合新闻类型 \texttt{\{news\_type\}} 和核心事件 \texttt{\{main\_event\}}，用简洁客观的语言交代最重要的信息。

3）基本情境：交代事件发生或报道涉及的时间 \texttt{\{time\}}、地点 \texttt{\{place\}} 和相关背景 \texttt{\{background\}}；如果某项信息为空，不要自行补充。

4）相关主体：围绕涉及的人物或机构 \texttt{\{people\_or\_orgs\}} 展开叙述，说明他们与事件之间的关系，不要添加资料之外的新主体。

5）事件经过：根据 \texttt{\{event\_process\}} 展开主体内容，使事件发展、前后因果和关键节点清楚连贯。

6）结果进展：根据 \texttt{\{result\_or\_status\}} 说明事件结果、当前状态或后续进展，避免推断资料中没有给出的未来发展。

7）来源补充：结尾可结合 \texttt{\{source\_or\_reporter\}} 和 \texttt{\{commentary\}} 补充消息来源、记者信息或评论性内容；如果这些信息为空，则围绕事件结果自然收束全文。

\medskip
\hrule
\medskip

\textbf{Level 3+Ref}

你是一名擅长中文新闻报道写作的专业写作者。请根据以下任务大纲生成一篇新闻报道。正文应信息清楚、叙述完整、语言客观。只输出正文，文本长度尽量接近\texttt{\{target\_length\}}。

任务大纲同Level 3。

下面提供一篇同领域的参考文本，注意仅用作参考。正文的主要信息、内容安排和写作重点应以任务大纲为准。

参考文本：\texttt{\{reference\_text\}}

\end{promptbox}
\end{CJK*}

This section presents the prompts used for the four text-production settings. The templates control the input information, generation scope, and content constraints, thereby defining different levels of machine involvement.

\noindent\textbf{Machine-Generated Text.}
Only the opening segment of a human-written text and its original length are provided. The complete source text, topic description, and detailed writing instructions are omitted. The model continues the text from the given prefix while keeping the final length close to that of the source. This minimal prompt leaves content planning and expression largely to the model.

\noindent\textbf{Content Filling.}
Within each domain, we divide the 1,000 source texts into three approximately equal groups, duplicate the groups, and assign the six resulting groups to the six generators, one group per generator. Each source text is thus processed by two generators under all four filling conditions. Separate prompts handle contiguous and dispersed spans. For contiguous spans, the model fills one mask from its preceding context while preserving the unmasked text and meeting the specified length constraint. For dispersed spans, it fills masks sequentially, receiving the updated text and current mask identifier at each step and generating only the targeted span. This avoids inconsistencies from filling multiple spans simultaneously and preserves coherence across generated segments.

\noindent\textbf{Machine Rewriting.}
Machine Rewriting uses separate prompts for Light Rewriting and Deep Rewriting. Light Rewriting permits local changes to wording, word order, and sentence transitions while preserving the original facts, meaning, paragraph structure, and overall style. Deep Rewriting additionally permits changes to sentence structure and within-paragraph organization, while retaining the topic, main facts, and domain-specific style. Both prompts require the model to output only the rewritten text and prohibit the introduction of new facts.

\noindent\textbf{Prompt-based Generation.}
Prompt-based Generation progressively increases the amount of structured information and writing constraints. Level 1 provides only the topic and target length. Level 2 adds key information, such as the main event and its outcome or current status. Level 3 further specifies the writing role, narrative organization, time, location, involved entities, event process, and source information. Level 3+Ref. additionally provides a same-domain reference text, but excludes the original source text of the current sample from reference selection. The reference guides content organization and domain-specific expression, while the generated text must follow the structured information of the current sample. News-domain templates are shown as examples; the other domains follow the same four-level hierarchy, with prompt content adapted to each domain.

\subsection{Dataset Statistics}
\label{app:dataset-statistics}

\begin{center}
\footnotesize

\refstepcounter{table}
\label{tab:hwt-length-statistics}

\raggedright
Table~\thetable: Length statistics of human-written texts across five domains. Max/Min, Mean, SD, CV, and Count denote maximum/minimum character length, mean character length, standard deviation, coefficient of variation, and sample count, respectively.

\par\smallskip

\setlength{\tabcolsep}{6pt}
\renewcommand{\arraystretch}{1.0}

\centering
\begin{tabular}{lcccccc}
\toprule
\textbf{Domain}
& \textbf{Max}
& \textbf{Min}
& \textbf{Mean}
& \textbf{SD}
& \textbf{CV}
& \textbf{Count} \\
\midrule
News       & 1707 & 1013 & 1296.22 & 157.88 & 12.18\% & 1000 \\
Law        & 1500 & 1000 & 1254.71 & 146.36 & 11.66\% & 1000 \\
Q\&A       & 1487 & 960  & 1223.92 & 143.98 & 11.76\% & 1000 \\
Encyclopedia & 1532 & 988 & 1250.24 & 147.30 & 11.78\% & 1000 \\
Literature & 1505 & 1000 & 1249.82 & 144.56 & 11.57\% & 1000 \\
\bottomrule
\end{tabular}
\end{center}
\begingroup
\small

\captionof{table}{Length statistics of machine-involved samples across collaborative task conditions. Max/Min, Mean, SD, CV, and Count denote maximum/minimum character length, mean character length, standard deviation, coefficient of variation, and sample count, respectively.}
\label{tab:app-task-length-statistics}

\begin{center}
\setlength{\tabcolsep}{4pt}
\renewcommand{\arraystretch}{1.08}

\begin{tabular}{lrrrrrrr}
\toprule
\textbf{Task}
& \textbf{Condition}
& \textbf{Max}
& \textbf{Min}
& \textbf{Mean}
& \textbf{SD}
& \textbf{CV}
& \textbf{Count} \\
\midrule

AI Text
& Overall
& 2062
& 611
& 1143.96
& 245.67
& 21.48\%
& 28321 \\

\midrule

Content Filling
& LC-C
& 1977
& 809
& 1274.13
& 178.74
& 14.03\%
& 9855 \\

Content Filling
& LC-D
& 1998
& 704
& 1188.14
& 196.41
& 16.53\%
& 9512 \\

Content Filling
& HC-C
& 2108
& 669
& 1220.80
& 219.61
& 17.99\%
& 9336 \\

Content Filling
& HC-D
& 2092
& 646
& 1111.41
& 225.42
& 20.28\%
& 8204 \\

\textbf{Content Filling}
& \textbf{Overall}
& \textbf{2108}
& \textbf{646}
& \textbf{1202.31}
& \textbf{212.89}
& \textbf{17.71\%}
& \textbf{36907} \\

\midrule

Machine Rewriting
& Light
& 1925
& 733
& 1219.49
& 158.70
& 13.01\%
& 29647 \\

Machine Rewriting
& Substantive
& 2106
& 616
& 1217.22
& 197.43
& 16.22\%
& 29391 \\

\textbf{Machine Rewriting}
& \textbf{Overall}
& \textbf{2106}
& \textbf{616}
& \textbf{1218.36}
& \textbf{179.04}
& \textbf{14.69\%}
& \textbf{59038} \\

\midrule

Prompt-based Generation
& Level 1
& 2133
& 625
& 1250.94
& 255.40
& 20.42\%
& 28447 \\

Prompt-based Generation
& Level 2
& 2124
& 638
& 1343.53
& 204.37
& 15.21\%
& 29341 \\

Prompt-based Generation
& Level 3
& 2096
& 625
& 1215.14
& 222.56
& 18.32\%
& 29369 \\

Prompt-based Generation
& Level 3+Ref
& 2101
& 611
& 1191.56
& 222.93
& 18.71\%
& 29116 \\

\textbf{Prompt-based Generation}
& \textbf{Overall}
& \textbf{2133}
& \textbf{611}
& \textbf{1250.39}
& \textbf{234.14}
& \textbf{18.73\%}
& \textbf{116273} \\

\bottomrule
\end{tabular}
\end{center}

\endgroup

\par\medskip

\noindent\textbf{The dataset combines balanced human-written sources with task-dependent variation in machine-involved samples.}

\smallskip
\noindent\textbf{Human-written sources.} Table~\ref{tab:hwt-length-statistics} contains 1,000 texts per domain. Their mean lengths range from 1,223.92 to 1,296.22 characters, with coefficients of variation between 11.57\% and 12.18\%.

\smallskip
\noindent\textbf{Task-level variation.} Table~\ref{tab:app-task-length-statistics} shows that mean sample length ranges from 1,143.96 characters for MGT to 1,250.39 for Prompt-based Generation. The latter also contributes the most samples (116,273), indicating unequal task sizes.

\smallskip
\noindent\textbf{Domain and generator coverage.} The task--domain--generator breakdown shows that every production mode spans all five domains and six generators. Total sample counts per generator range from 37,957 (Kimi) to 41,268 (GPT).

\par\medskip


\par\medskip

\refstepcounter{table}
\label{tab:app-a-machine-text-statistics}

\noindent Table~\thetable:
Counts of machine-involved samples by production type, domain, and generative model. GPT, Qwen, GLM, Kimi, DeepSeek, and Gemini
denote GPT-5.5, Qwen3.7-Plus, GLM-5.2, Kimi K2.6, DeepSeek V4 Pro, and
Gemini 3.5 Flash.

\begin{center}
\footnotesize
\setlength{\tabcolsep}{3pt}
\renewcommand{\arraystretch}{1.08}

\setlength{\heavyrulewidth}{0.08em}
\setlength{\lightrulewidth}{0.05em}
\setlength{\aboverulesep}{0.45ex}
\setlength{\belowrulesep}{0.45ex}

\begin{tabular}{ccccccccc}
\toprule
\textbf{Task}
& \textbf{Domain}
& \textbf{GPT}
& \textbf{Qwen}
& \textbf{GLM}
& \textbf{Kimi}
& \textbf{DeepSeek}
& \textbf{Gemini}
& \textbf{Total} \\
\midrule

\multirow[c]{5}{*}{MGT}
& News         & 992 & 977 & 907 & 943 & 933 & 955 & 5707 \\
& Law          & 999 & 978 & 918 & 828 & 863 & 991 & 5577 \\
& Q\&A         & 996 & 989 & 953 & 937 & 872 & 984 & 5731 \\
& Encyclopedia & 973 & 964 & 894 & 944 & 939 & 971 & 5685 \\
& Literature   & 950 & 976 & 952 & 914 & 852 & 977 & 5621 \\
\midrule

\textbf{MGT Total}
& & 4910 & 4884 & 4624 & 4566 & 4459 & 4878 & \textbf{28321} \\
\midrule

\multirow[c]{5}{*}{Content Filling}
& News         & 1320 & 1236 & 1234 & 1196 & 1237 & 1312 & 7535 \\
& Law          & 1329 & 1208 & 1260 & 1156 & 1305 & 1314 & 7572 \\
& Q\&A         & 1319 & 1227 & 1273 & 1067 & 1210 & 1302 & 7398 \\
& Encyclopedia & 1310 & 1206 & 1260 & 1162 & 1223 & 1291 & 7452 \\
& Literature   & 1298 & 1252 & 1208 & 678  & 1240 & 1274 & 6950 \\
\midrule

\textbf{Content Infill Total}
& & 6576 & 6129 & 6235 & 5259 & 6215 & 6493 & \textbf{36907} \\
\midrule

\multirow[c]{5}{*}{Machine Rewriting}
& News         & 1940 & 1922 & 1930 & 1917 & 1948 & 1911 & 11568 \\
& Law          & 2000 & 2000 & 2000 & 2000 & 1999 & 1996 & 11995 \\
& Q\&A         & 1968 & 1996 & 1988 & 1969 & 1994 & 1984 & 11899 \\
& Encyclopedia & 1996 & 1925 & 1926 & 1937 & 1926 & 1958 & 11668 \\
& Literature   & 1993 & 1987 & 1991 & 1993 & 1985 & 1959 & 11908 \\
\midrule

\textbf{Machine Rewriting Total}
& & 9897 & 9830 & 9835 & 9816 & 9852 & 9808 & \textbf{59038} \\
\midrule

\multirow[c]{5}{*}{Prompt-based Generation}
& News         & 4000 & 3962 & 3938 & 3207 & 3837 & 3908 & 22852 \\
& Law          & 3978 & 3948 & 3982 & 3827 & 3659 & 3895 & 23289 \\
& Q\&A         & 3986 & 3957 & 3973 & 3935 & 3972 & 3978 & 23801 \\
& Encyclopedia & 3985 & 3794 & 3925 & 3885 & 3911 & 3975 & 23475 \\
& Literature   & 3936 & 3979 & 3973 & 3462 & 3558 & 3948 & 22856 \\
\midrule

\textbf{Prompt-based Generation Total}
& & 19885 & 19640 & 19791 & 18316 & 18937 & 19704 & \textbf{116273} \\
\midrule

\textbf{Total}
& & 41268 & 40483 & 40485 & 37957 & 39463 & 40883 & \textbf{240539} \\
\bottomrule
\end{tabular}
\end{center}

\par\medskip

\section{Detector Details}
\label{app:detector-details}

\noindent\textbf{Zero-Shot Detectors}\par

\noindent\textbf{Log-Likelihood}~\citep{gehrmann2019gltr} uses the mean log-likelihood assigned by a reference language model to the observed tokens as its detection score. The underlying assumption is that machine-generated text tends to conform more closely to the probability distribution learned by the generator and may therefore receive a higher likelihood.

\noindent\textbf{LogRank}~\citep{gehrmann2019gltr} performs detection using the average rank of the observed tokens in the language model's predictive distributions. Because generation procedures tend to select highly ranked candidate tokens, their rank distributions may differ from those of human-written text.

\noindent\textbf{Entropy}~\citep{gehrmann2019gltr} measures the predictive uncertainty of a language model at each token position and aggregates these values into a document-level score. This signal captures differences in the concentration of the model's predictive distributions between human-written and machine-generated texts.

\noindent\textbf{DetectLLM-LRR}~\citep{su2023detectllm} combines text likelihood with token log-rank through a likelihood--log-rank ratio. This criterion jointly captures the model's absolute predictive confidence and the relative positions of the observed tokens within its candidate distributions.

\noindent\textbf{LastDE}~\citep{xu2024lastde} treats the token-probability sequence as a time series and uses multiscale diversity entropy to characterize its local dynamics. These features are combined with global likelihood to distinguish the probability variation patterns of human-written and machine-generated texts.

\noindent\textbf{LastDE++}~\citep{xu2024lastde} augments these dynamic features with efficient sampling and score normalization. Its final criterion compares the score of the observed text with those of contrast samples, improving the stability of the resulting predictions.

\noindent\textbf{SpecDetect}~\citep{luo2025specdetect} analyzes token log-probability sequences in the frequency domain and uses the total spectral energy obtained through the discrete Fourier transform as its detection signal. This feature reflects differences in the magnitude and frequency structure of probability fluctuations.

\noindent\textbf{SpecDetect++}~\citep{luo2025specdetect} further introduces a sampling-based discrepancy mechanism. It compares the spectral energy of the observed text with the distribution derived from model-generated contrast samples and uses the standardized difference as the detection score.

\noindent\textbf{Fast-DetectGPT}~\citep{bao2024fastdetectgpt} relies on conditional probability curvature, comparing the conditional probabilities of observed tokens with those of model-sampled alternatives. It thereby approximates the curvature-based criterion of DetectGPT at substantially lower computational cost.

\noindent\textbf{Binoculars}~\citep{hans2024binoculars} uses an observer model and a performer model to compute perplexity and cross-perplexity, respectively, and determines text authorship from their ratio. Detection is therefore based on the relative confidence exhibited by two language models on the same text.

\noindent\textbf{Supervised Detectors}\par

\noindent\textbf{RADAR}~\citep{hu2023radar} improves detection robustness through adversarial training between a paraphraser and a detector. The paraphraser rewrites machine-generated text to evade detection, while the detector learns to identify both the original and rewritten outputs. We directly use the publicly released detector checkpoint.

\noindent\textbf{ReMoDetect}~\citep{lee2024remodetect} builds on reward modeling and employs experience replay to mitigate overfitting during continual preference fine-tuning. It further uses mixed-preference data created through partial machine rewriting to refine the classification boundary. We perform inference with its publicly released DeBERTa-based checkpoint.

\noindent\textbf{Longformer}~\citep{li2024mage} denotes the long-document detector released with MAGE. It is trained on human-written texts and machine-generated texts from multiple generators. We use the publicly available checkpoint.

\noindent\textbf{BLOOMZ-3B}~\citep{muennighoff2023crosslingual} is a binary classifier built on the multilingual instruction-tuned BLOOMZ-3B model and trained on a mixture of human-written and machine-generated texts. We use the publicly released mixed-detector checkpoint.

\noindent\textbf{OpenAI RoBERTa}~\citep{solaiman2019release} predicts text authorship using a RoBERTa classifier fine-tuned on human-written WebText documents and GPT-2 outputs. Although it was trained primarily on outputs from an earlier generation of language models, it provides a useful baseline for evaluating cross-model and cross-lingual transfer.

\noindent\textbf{ConDA}~\citep{bhattacharjee2023conda} combines contrastive learning with unsupervised domain adaptation. Using labeled source-domain data and unlabeled target-domain data, it learns representations that are both class-discriminative and domain-invariant. We use its publicly released RoBERTa-based detector checkpoint.

\noindent\textbf{ChatGPT Detector}~\citep{guo2023hc3} is a RoBERTa classifier fine-tuned on paired human and ChatGPT answers from HC3. We adopt its publicly released Chinese checkpoint to examine how a detector trained on Chinese question-answering data transfers to collaborative texts from multiple domains.

\noindent\textbf{AIGC-detector-zhv2}~\citep{tian2024multiscale} is built on Chinese RoBERTa-wwm-ext and trained using a multiscale positive--unlabeled learning framework for Chinese machine-generated text detection. In the cross-condition experiments, we initialize the detector with its public weights and train it separately on the data from each source condition.

\noindent\textbf{IMBD}~\citep{chen2025imitate} first applies Style Preference Optimization to align a scoring model with machine-writing style preferences and then derives its detection score from Style-Conditional Probability Curvature. Following the original procedure, we train the model separately under each source condition to evaluate transfer across domains, collaborative tasks, and generative models.

\noindent\textbf{Boundary Localization Detectors}\par

\noindent\textbf{WCP}~\citep{li2026segmenting} approaches boundary localization in human--AI mixed texts from a change-point detection perspective. It converts each sentence into a scalar detection score, forming a one-dimensional sequence, and then applies a weighted change-point detection algorithm to identify shifts in the score distribution. We use its weighted variant for sentence-level boundary prediction.

\noindent\textbf{DeBERTa--BiGRU--CRF.}
We implement a token-level detector with a DeBERTa-v3-base encoder,
a three-layer bidirectional GRU, and a CRF decoder to assign
human- or machine-authorship labels to each token.
A DeBERTa--BiGRU--CRF configuration has also been evaluated in
DAMASHA~\citep{saiteja2026damasha}.
We merge consecutive machine-labeled tokens into character-level spans
and map them to the shared sentence-level representation for evaluation.

\section{Experimental Setup and Data Preprocessing}
\label{app:experimenta-setup}

\begin{figure}[h]
    \centering
    \includegraphics[width=0.95\textwidth]{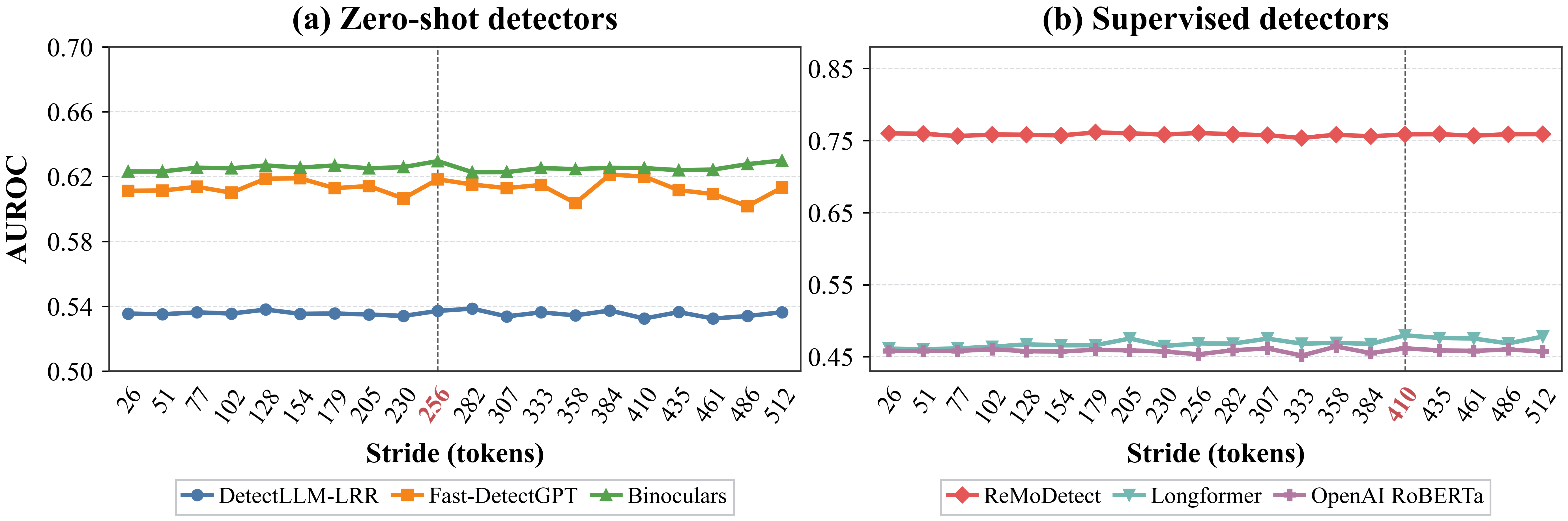}
    \caption{stride-sensitivity analysis for representative zero-shot and supervised detectors. The left panel shows zero-shot detectors, and the right panel shows supervised detectors. The highlighted x-axis value indicates the selected stride for the corresponding experiment.}
    \label{fig:striding-window}
    \vspace{-10pt}
\end{figure}

\subsection{Sliding-Window Strategy and Stride Selection}
\label{app:sliding-window}

\noindent\textbf{Windowing strategy.}
Each text is tokenized with the corresponding detector's tokenizer and divided into 512-token windows with stride \(s\). If the regular windows leave the text ending uncovered, we include an additional end-aligned window. Windows are scored independently, and their scores are averaged to obtain the document-level score.

\noindent\textbf{Stride sensitivity.}
We use 973 GPT-generated texts from the News domain of the Machine Rewriting task and their 973 matched human-written sources. With 512-token windows, we vary the overlap ratio from \(0\%\) to \(95\%\) in increments of five percentage points and compare the resulting strides using document-level AUROC.

\noindent\textbf{Stride selection.}
Figure~\ref{fig:striding-window} shows that mean AUROC peaks at stride 256 across DetectLLM-LRR~\citep{su2023detectllm}, Fast-DetectGPT~\citep{bao2024fastdetectgpt}, and Binoculars~\citep{hans2024binoculars} (left). For ReMoDetect~\citep{lee2024remodetect}, Longformer~\citep{beltagy2020longformer}, and OpenAI RoBERTa~\citep{solaiman2019release}, the mean peaks at stride 410 (right). We therefore adopt 256- and 410-token strides for zero-shot and supervised detectors, respectively. AUROC varies little across adjacent strides in this setting.

\subsection{Boundary Localization Data Processing}

\noindent\textbf{We construct sentence-level boundary-localization instances by deriving gold machine-generated intervals, aligning detector outputs to the sentence level, and evaluating all methods under source-disjoint data splits.}

\noindent\textbf{Gold interval construction.} We use the content-filling subset from Experiment~1. For each sample, the original text containing \texttt{[MASK]} markers is aligned with its corresponding generated text, and the spans replacing the masks are treated as gold machine-generated intervals. Samples with unreliable alignment are excluded.

\noindent\textbf{Sentence-level annotation.} Each text is segmented according to Chinese and English sentence-final punctuation. A sentence is labeled as machine-generated if it overlaps with at least one gold interval; otherwise, it is labeled as human-written. The gold boundaries are defined at positions where the labels of adjacent sentences change.

\noindent\textbf{Prediction alignment.} To place all detector outputs at the same granularity, WCP predicts change points directly over sentence sequences. The token-level predictions produced by DeBERTa–BiGRU–CRF are merged into character-level intervals and then mapped to sentence labels and boundary positions.

\noindent\textbf{Data splitting and evaluation.} Samples are grouped by domain and source-text ID before being divided into training, validation, and test sets, ensuring that variants derived from the same source text do not appear across different splits. We conduct the final evaluation on the test set using Cohen's $\kappa$, WindowDiff, and Boundary $F_1@1$.

\subsection{Cross-Condition Generalization Data Processing}

\noindent\textbf{Cross-condition generalization.} Experiment~4 evaluates how supervised detectors transfer when the training and test conditions differ.

\begin{itemize}[
    leftmargin=1.2em,
    labelindent=0pt,
    labelwidth=0.7em,
    labelsep=0.5em,
    itemsep=2pt,
    topsep=2pt
]
    \item \textit{Generalization axes.} We construct train--test matrices separately along the domain, collaborative-task, and generative-model dimensions, resulting in $5\times5$, $4\times4$, and $6\times6$ matrices, respectively.

    \item \textit{Training and labels.} For each matrix row, one condition is used for training and every condition along the same dimension is used for testing. Detectors are trained on the corresponding training split and checkpointed using the validation split from that condition. Human-written texts form the negative class, whereas machine-involved texts form the positive class.

    \item \textit{Threshold selection.} For each train--test cell, we select the decision threshold that maximizes Youden's $J$ on its labeled evaluation samples.

    \item \textit{Evaluation.} We report Accuracy at the cell-specific Youden threshold. Scores are macro-averaged after source-ID pairing and fine-grained condition grouping to reduce the influence of unequal sample counts across matrix cells.
\end{itemize}

\section{Supplementary Experimental Results}
\label{app:suplementary-experimental}

\subsection{Detection Performance at Fixed False-Positive Rates}
\label{app:fine-grained-results}


\definecolor{fprpureai}{HTML}{F6E4E4}
\definecolor{fprcontentfill}{HTML}{E7EEF8}
\definecolor{fprrewriting}{HTML}{E8F3E8}
\definecolor{fprpromptgen}{HTML}{FFF1D6}
\definecolor{fprbestresult}{HTML}{FFD966}
\definecolor{fprsecondresult}{HTML}{D9D9D9}

\newcommand{\fprbest}[1]{\cellcolor{fprbestresult}#1}
\newcommand{\fprsecond}[1]{\cellcolor{fprsecondresult}#1}
\newcommand{\fprfirstmark}{\colorbox{fprbestresult}{\strut 1st}}
\newcommand{\fprsecondmark}{\colorbox{fprsecondresult}{\strut 2nd}}

\begin{table}[!tp]
\centering
\caption{Detector performance under human--AI collaboration, measured by ACC and F1 at 1\%, 5\%, and 10\% FPR. The best two results are highlighted by \fprfirstmark{} and \fprsecondmark{}. Values are reported as percentages and macro-averaged over 30 domain--generation-model combinations.}
\label{tab:finegrained_acc_f1_all_fpr}

\begingroup
\fontsize{7}{8.2}\selectfont
\setlength{\tabcolsep}{0.8pt}
\renewcommand{\arraystretch}{1.0}

\begin{tabular*}{\linewidth}{
@{\extracolsep{\fill}}
l
*{2}{>{\columncolor{fprpureai}\centering\arraybackslash}c}
*{8}{>{\columncolor{fprcontentfill}\centering\arraybackslash}c}
*{4}{>{\columncolor{fprrewriting}\centering\arraybackslash}c}
*{8}{>{\columncolor{fprpromptgen}\centering\arraybackslash}c}
@{}
}

\toprule

\rowcolor{white}
\multirow{3}{*}{\textbf{Detector}}
& \multicolumn{2}{c}{\textbf{MGT}}
& \multicolumn{8}{c}{\textbf{Content Filling}}
& \multicolumn{4}{c}{\textbf{Machine Rewriting}}
& \multicolumn{8}{c}{\textbf{Prompt-based Generation}} \\

\cmidrule(lr){2-3}
\cmidrule(lr){4-11}
\cmidrule(lr){12-15}
\cmidrule(lr){16-23}

\rowcolor{white}
& \multicolumn{2}{c}{}
& \multicolumn{2}{c}{LC-C}
& \multicolumn{2}{c}{LC-D}
& \multicolumn{2}{c}{HC-C}
& \multicolumn{2}{c}{HC-D}
& \multicolumn{2}{c}{Light}
& \multicolumn{2}{c}{Subst.}
& \multicolumn{2}{c}{L1}
& \multicolumn{2}{c}{L2}
& \multicolumn{2}{c}{L3}
& \multicolumn{2}{c}{L3+Ref.} \\

\cmidrule(lr){2-3}
\cmidrule(lr){4-11}
\cmidrule(lr){12-15}
\cmidrule(lr){16-23}

\rowcolor{white}
& \textbf{ACC} & \textbf{F1}
& \textbf{ACC} & \textbf{F1}
& \textbf{ACC} & \textbf{F1}
& \textbf{ACC} & \textbf{F1}
& \textbf{ACC} & \textbf{F1}
& \textbf{ACC} & \textbf{F1}
& \textbf{ACC} & \textbf{F1}
& \textbf{ACC} & \textbf{F1}
& \textbf{ACC} & \textbf{F1}
& \textbf{ACC} & \textbf{F1}
& \textbf{ACC} & \textbf{F1} \\

\midrule

\multicolumn{23}{@{}l@{}}{\textbf{1\% FPR}\quad\textit{Zero-shot detectors}} \\
\midrule

Log-Likelihood
& 58.5 & 22.5
& \fprsecond{52.6} & \fprsecond{10.4}
& \fprsecond{52.3} & \fprsecond{9.6}
& 53.8 & 14.1
& \fprsecond{54.8} & \fprsecond{15.7}
& \fprsecond{52.5} & \fprsecond{10.0}
& \fprsecond{52.6} & \fprsecond{10.3}
& 56.7 & 18.9
& 53.4 & 12.2
& 53.6 & 12.7
& 53.9 & 13.5 \\

LogRank
& 58.4 & 21.5
& 52.4 & 9.6
& \fprsecond{52.3} & 9.1
& 53.4 & 12.8
& 54.5 & 14.8
& 52.2 & 9.0
& 52.4 & 9.4
& 56.9 & 18.5
& 53.2 & 11.4
& 53.4 & 11.8
& 53.7 & 12.4 \\

Entropy
& 50.1 & 2.3
& 49.9 & 1.7
& 49.9 & 1.6
& 49.9 & 1.4
& 50.0 & 1.7
& 49.9 & 1.4
& 50.0 & 1.7
& 50.0 & 1.7
& 49.8 & 1.1
& 50.0 & 1.7
& 50.0 & 1.6 \\

DetectLLM-LRR
& 53.2 & 10.7
& 50.4 & 3.2
& 50.4 & 3.3
& 50.6 & 3.8
& 50.6 & 4.0
& 50.2 & 2.5
& 50.2 & 2.5
& 53.4 & 10.8
& 50.6 & 3.7
& 50.7 & 4.3
& 50.8 & 4.4 \\

Fast-DetectGPT
& 54.9 & 18.5
& 51.5 & 7.4
& 51.2 & 6.4
& 52.1 & 9.6
& 52.1 & 9.0
& 50.7 & 4.7
& 50.2 & 2.9
& 54.1 & 15.7
& 51.0 & 5.5
& 50.8 & 4.8
& 50.9 & 5.2 \\

Binoculars
& \fprbest{69.2} & \fprbest{52.3}
& 51.5 & 7.4
& 50.7 & 4.6
& \fprsecond{58.1} & \fprsecond{26.7}
& 53.7 & 14.8
& 50.8 & 5.1
& 51.0 & 5.5
& \fprbest{65.4} & \fprbest{44.6}
& \fprsecond{58.1} & \fprsecond{26.3}
& \fprsecond{55.1} & \fprsecond{18.2}
& \fprsecond{55.4} & \fprsecond{19.4} \\

LastDE
& 53.9 & 14.9
& 50.5 & 3.7
& 50.4 & 3.6
& 50.9 & 5.1
& 51.1 & 6.1
& 50.4 & 3.2
& 50.8 & 4.6
& 54.1 & 14.7
& 52.7 & 10.6
& 51.7 & 7.6
& 51.8 & 7.9 \\

LastDE++
& 60.0 & 32.0
& 51.5 & 7.6
& 51.0 & 5.9
& 54.4 & 17.1
& 52.5 & 11.2
& 50.9 & 5.2
& 50.8 & 4.7
& 57.3 & 25.0
& 53.8 & 14.0
& 52.0 & 9.1
& 52.0 & 9.1 \\

SpecDetect
& \fprsecond{66.3} & \fprsecond{45.4}
& \fprbest{56.4} & \fprbest{21.7}
& \fprbest{55.8} & \fprbest{19.9}
& \fprbest{60.5} & \fprbest{33.0}
& \fprbest{59.8} & \fprbest{29.9}
& \fprbest{56.8} & \fprbest{22.1}
& \fprbest{57.2} & \fprbest{23.2}
& \fprsecond{65.1} & \fprsecond{42.5}
& \fprbest{60.3} & \fprbest{32.4}
& \fprbest{58.4} & \fprbest{27.1}
& \fprbest{58.6} & \fprbest{27.6} \\

SpecDetect++
& 61.9 & 37.2
& 52.0 & 9.4
& 51.5 & 7.6
& 55.2 & 19.6
& 53.3 & 13.8
& 51.7 & 8.2
& 51.5 & 7.5
& 59.7 & 32.0
& 55.5 & 19.7
& 53.2 & 13.3
& 53.3 & 13.6 \\

\midrule

\multicolumn{23}{@{}l@{}}{\textbf{1\% FPR}\quad\textit{Supervised detectors}} \\
\midrule

RADAR
& 49.9 & 1.4
& 49.7 & 0.6
& 49.7 & 1.0
& 49.6 & 0.4
& 49.6 & 0.5
& 49.6 & 0.4
& 49.6 & 0.3
& 49.6 & 0.5
& 49.6 & 0.2
& 49.5 & 0.0
& 49.5 & 0.0 \\

ReMoDetect
& \fprsecond{59.5} & \fprsecond{28.4}
& \fprsecond{52.2} & \fprsecond{9.6}
& \fprsecond{51.1} & \fprsecond{6.0}
& \fprbest{57.1} & \fprbest{23.2}
& \fprbest{53.6} & \fprbest{13.8}
& \fprbest{51.9} & \fprbest{8.7}
& \fprbest{53.9} & \fprbest{14.8}
& \fprsecond{60.2} & \fprsecond{29.3}
& \fprbest{60.5} & \fprbest{28.9}
& \fprbest{59.5} & \fprbest{26.3}
& \fprbest{59.4} & \fprbest{26.2} \\

Longformer
& 55.6 & 19.9
& 49.8 & 1.4
& 50.0 & 1.8
& 49.8 & 1.3
& 49.8 & 1.3
& 49.8 & 1.0
& 50.1 & 2.4
& 56.7 & 23.0
& 51.4 & 6.6
& 50.7 & 4.6
& 50.8 & 4.8 \\

BLOOMZ-3B
& \fprbest{92.8} & \fprbest{90.4}
& \fprbest{52.7} & \fprbest{11.4}
& \fprbest{51.9} & \fprbest{8.8}
& \fprsecond{52.8} & \fprsecond{11.8}
& \fprsecond{52.5} & \fprsecond{10.8}
& \fprsecond{50.6} & \fprsecond{4.3}
& \fprsecond{51.3} & \fprsecond{6.7}
& \fprbest{98.2} & \fprbest{98.0}
& \fprsecond{57.2} & \fprsecond{24.7}
& \fprsecond{56.9} & \fprsecond{22.4}
& \fprsecond{57.4} & \fprsecond{23.3} \\

OpenAI RoBERTa
& 49.8 & 1.1
& 49.8 & 1.2
& 49.9 & 1.5
& 49.8 & 1.1
& 49.7 & 0.8
& 50.1 & 2.3
& 50.2 & 2.5
& 49.9 & 1.3
& 49.7 & 0.8
& 49.7 & 0.7
& 49.7 & 0.8 \\

ConDA
& 49.8 & 1.1
& 49.8 & 1.3
& 49.9 & 1.5
& 49.8 & 1.1
& 49.8 & 1.1
& 49.8 & 1.0
& 49.8 & 0.9
& 49.9 & 1.4
& 49.9 & 1.6
& 49.8 & 1.2
& 49.8 & 1.2 \\

ChatGPT Detector
& 51.9 & 8.3
& 50.3 & 2.8
& 50.2 & 2.5
& 50.9 & 4.9
& 50.7 & 4.2
& 50.4 & 3.1
& 50.3 & 2.7
& 51.2 & 6.1
& 50.6 & 4.0
& 51.2 & 6.0
& 51.4 & 6.4 \\

\midrule[0.6pt]

\multicolumn{23}{@{}l@{}}{\textbf{5\% FPR}\quad\textit{Zero-shot detectors}} \\
\midrule

Log-Likelihood
& 60.8 & 30.3
& 56.1 & 22.2
& \fprsecond{55.7} & 21.2
& 58.3 & 27.6
& 58.4 & 26.5
& 56.3 & 22.1
& \fprsecond{56.5} & 22.4
& 60.1 & 28.4
& 57.8 & 24.6
& 57.9 & 24.7
& 58.1 & 25.1 \\

LogRank
& 60.0 & 27.9
& 55.3 & 19.8
& 55.0 & 18.8
& 57.2 & 24.5
& 57.2 & 23.3
& 55.2 & 19.2
& 55.5 & 19.4
& 59.2 & 25.7
& 56.7 & 21.4
& 56.7 & 21.1
& 57.0 & 21.7 \\

Entropy
& 49.2 & 5.6
& 49.2 & 5.7
& 49.3 & 6.1
& 48.8 & 4.5
& 49.2 & 5.2
& 49.3 & 6.0
& 49.8 & 7.4
& 49.0 & 4.6
& 48.8 & 4.0
& 49.3 & 5.4
& 49.2 & 5.3 \\

DetectLLM-LRR
& 55.3 & 17.4
& 51.1 & 9.5
& 51.1 & 9.6
& 51.6 & 11.0
& 51.8 & 11.1
& 50.5 & 8.3
& 50.6 & 8.4
& 55.2 & 16.4
& 51.5 & 10.2
& 51.6 & 10.2
& 51.7 & 10.5 \\

Fast-DetectGPT
& 60.5 & 37.9
& 54.4 & 22.1
& 53.7 & 20.4
& 56.0 & 26.6
& 55.0 & 23.0
& 52.5 & 16.6
& 51.0 & 11.9
& 59.8 & 35.3
& 53.1 & 17.9
& 52.5 & 16.7
& 52.9 & 17.7 \\

Binoculars
& \fprbest{76.9} & \fprbest{69.5}
& 54.0 & 21.4
& 52.3 & 16.2
& \fprsecond{64.2} & 45.6
& 58.3 & 32.1
& 52.9 & 18.3
& 53.5 & 19.9
& \fprsecond{74.1} & \fprsecond{65.1}
& \fprsecond{65.3} & \fprsecond{48.4}
& \fprsecond{60.9} & \fprsecond{38.6}
& \fprsecond{61.5} & \fprsecond{39.9} \\

LastDE
& 61.4 & 38.4
& 52.2 & 15.3
& 52.0 & 14.4
& 54.1 & 20.8
& 55.1 & 23.6
& 52.2 & 15.2
& 54.0 & 20.1
& 61.8 & 38.1
& 59.7 & 32.6
& 56.8 & 25.5
& 56.9 & 25.7 \\

LastDE++
& 69.7 & 57.6
& 55.3 & 25.2
& 53.7 & 20.7
& 61.6 & 41.0
& 58.2 & 32.9
& 54.2 & 22.3
& 53.7 & 20.1
& 67.1 & 52.7
& 60.7 & 38.0
& 56.8 & 28.8
& 57.1 & 29.6 \\

SpecDetect
& \fprsecond{74.7} & \fprsecond{65.4}
& \fprbest{63.3} & \fprbest{41.4}
& \fprbest{61.5} & \fprbest{37.2}
& \fprbest{68.9} & \fprbest{54.7}
& \fprbest{67.1} & \fprbest{50.1}
& \fprbest{63.0} & \fprbest{40.0}
& \fprbest{63.4} & \fprbest{40.6}
& \fprbest{74.8} & \fprbest{65.5}
& \fprbest{70.9} & \fprbest{58.2}
& \fprbest{66.6} & \fprbest{48.6}
& \fprbest{66.9} & \fprbest{49.3} \\

SpecDetect++
& 72.4 & 62.9
& \fprsecond{57.2} & \fprsecond{30.5}
& 55.4 & \fprsecond{25.6}
& 63.7 & \fprsecond{46.0}
& \fprsecond{60.3} & \fprsecond{38.4}
& \fprsecond{56.5} & \fprsecond{28.7}
& 56.2 & \fprsecond{27.5}
& 71.7 & 61.3
& 64.8 & 47.5
& 60.6 & 38.2
& 61.0 & 39.1 \\

\midrule

\multicolumn{23}{@{}l@{}}{\textbf{5\% FPR}\quad\textit{Supervised detectors}} \\
\midrule

RADAR
& 49.9 & 8.1
& 48.8 & 4.3
& 49.3 & 6.1
& 48.0 & 1.9
& 48.4 & 3.4
& 48.5 & 3.4
& 48.5 & 3.6
& 50.2 & 8.8
& 48.1 & 2.2
& 47.7 & 0.9
& 47.8 & 1.0 \\

ReMoDetect
& \fprsecond{70.9} & \fprsecond{55.1}
& \fprbest{58.8} & \fprbest{31.7}
& \fprbest{56.7} & \fprbest{27.3}
& \fprbest{66.8} & \fprbest{48.0}
& \fprbest{62.8} & \fprbest{40.5}
& \fprbest{59.1} & \fprbest{33.0}
& \fprbest{63.6} & \fprbest{42.4}
& \fprsecond{71.5} & 55.5
& \fprbest{70.1} & \fprbest{49.8}
& \fprbest{69.2} & \fprbest{47.1}
& \fprbest{69.0} & \fprbest{46.8} \\

Longformer
& 67.3 & 51.7
& 50.2 & 9.6
& 50.1 & 9.0
& 50.4 & 10.2
& 50.1 & 9.4
& 49.9 & 8.4
& 50.7 & 11.1
& 71.3 & \fprsecond{59.8}
& 55.8 & 25.3
& 54.4 & 21.6
& 54.8 & 22.4 \\

BLOOMZ-3B
& \fprbest{91.6} & \fprbest{89.6}
& \fprsecond{57.5} & \fprsecond{29.4}
& \fprsecond{55.3} & \fprsecond{23.9}
& \fprsecond{58.2} & \fprsecond{30.9}
& \fprsecond{56.4} & \fprsecond{26.4}
& \fprsecond{51.6} & \fprsecond{13.8}
& \fprsecond{51.4} & \fprsecond{13.6}
& \fprbest{96.5} & \fprbest{96.6}
& \fprsecond{59.5} & \fprsecond{35.1}
& \fprsecond{58.4} & \fprsecond{31.4}
& \fprsecond{58.8} & \fprsecond{32.1} \\

OpenAI RoBERTa
& 48.9 & 4.9
& 49.1 & 5.2
& 49.3 & 6.0
& 48.9 & 4.6
& 48.5 & 3.7
& 50.4 & 9.4
& 50.7 & 10.4
& 49.0 & 5.1
& 48.5 & 3.4
& 48.4 & 3.2
& 48.5 & 3.6 \\

ConDA
& 49.7 & 7.8
& 49.7 & 7.6
& 49.7 & 7.5
& 49.6 & 7.5
& 48.9 & 5.3
& 48.7 & 4.5
& 48.7 & 4.3
& 50.4 & 9.8
& 51.7 & 14.0
& 50.3 & 9.6
& 50.2 & 9.4 \\

ChatGPT Detector
& 53.7 & 17.4
& 51.3 & 11.2
& 50.5 & 9.0
& 54.4 & 19.1
& 51.7 & 12.5
& 51.3 & 10.8
& 51.2 & 10.5
& 53.1 & 15.8
& 52.9 & 15.5
& 54.2 & 18.1
& 54.6 & 18.9 \\

\midrule[0.6pt]

\multicolumn{23}{@{}l@{}}{\textbf{10\% FPR}\quad\textit{Zero-shot detectors}} \\
\midrule

Log-Likelihood
& 60.1 & 33.3
& 56.8 & 27.1
& 56.2 & 25.6
& 58.9 & 32.2
& 58.6 & 30.6
& 56.7 & 26.5
& 57.1 & 27.0
& 59.6 & 32.1
& 58.1 & 28.7
& 58.1 & 28.6
& 58.4 & 29.3 \\

LogRank
& 59.2 & 30.8
& 55.8 & 24.2
& 55.4 & 23.2
& 57.9 & 29.3
& 57.4 & 27.1
& 55.7 & 23.5
& 55.8 & 23.5
& 58.6 & 29.0
& 56.9 & 25.5
& 56.9 & 25.1
& 57.1 & 25.5 \\

Entropy
& 48.1 & 8.9
& 48.5 & 10.5
& 48.8 & 11.2
& 47.6 & 7.6
& 48.2 & 8.9
& 48.8 & 11.3
& 49.6 & 13.1
& 47.8 & 7.4
& 47.5 & 6.6
& 48.5 & 9.1
& 48.5 & 9.1 \\

DetectLLM-LRR
& 55.5 & 21.4
& 51.7 & 14.3
& 51.5 & 13.7
& 52.2 & 15.9
& 52.0 & 14.9
& 50.8 & 12.6
& 50.9 & 12.5
& 55.0 & 19.4
& 52.0 & 14.4
& 51.9 & 13.8
& 51.9 & 14.0 \\

Fast-DetectGPT
& 63.9 & 49.1
& 56.5 & 33.1
& 55.8 & 31.5
& 58.5 & 37.8
& 57.0 & 33.2
& 53.9 & 26.4
& 51.6 & 19.9
& 63.1 & 46.7
& 55.0 & 28.5
& 54.3 & 27.2
& 54.6 & 28.1 \\

Binoculars
& \fprbest{79.8} & \fprbest{76.1}
& 56.4 & 33.6
& 53.8 & 26.7
& 68.0 & 57.1
& 60.8 & 42.4
& 55.2 & 30.4
& 56.3 & 32.9
& \fprsecond{77.9} & \fprsecond{73.5}
& 69.7 & 60.3
& 65.0 & 51.2
& 65.6 & 52.4 \\

LastDE
& 66.2 & 52.1
& 54.6 & 27.0
& 54.1 & 25.9
& 57.9 & 35.3
& 58.9 & 37.4
& 54.4 & 27.0
& 57.2 & 33.3
& 66.6 & 52.2
& 64.3 & 47.1
& 60.5 & 38.6
& 60.3 & 38.3 \\

LastDE++
& 74.9 & 69.4
& 59.0 & 39.9
& 57.0 & 35.2
& 66.9 & 56.1
& 62.7 & 48.0
& 58.0 & 37.9
& 57.0 & 34.6
& 73.4 & 67.2
& 66.2 & 54.2
& 61.6 & 44.8
& 62.0 & 45.9 \\

SpecDetect
& \fprsecond{77.6} & \fprsecond{72.7}
& \fprbest{65.7} & \fprbest{50.7}
& \fprbest{63.5} & \fprbest{45.7}
& \fprbest{71.7} & \fprbest{63.1}
& \fprbest{69.6} & \fprbest{58.6}
& \fprbest{64.7} & \fprbest{48.0}
& \fprbest{65.3} & \fprbest{48.6}
& \fprbest{78.4} & \fprbest{73.8}
& \fprbest{75.1} & \fprbest{68.2}
& \fprbest{70.2} & \fprbest{59.1}
& \fprbest{70.3} & \fprbest{59.4} \\

SpecDetect++
& 77.0 & 72.7
& \fprsecond{60.9} & \fprsecond{43.9}
& \fprsecond{58.5} & \fprsecond{38.5}
& \fprsecond{68.4} & \fprsecond{58.8}
& \fprsecond{64.7} & \fprsecond{52.0}
& \fprsecond{60.5} & \fprsecond{43.5}
& \fprsecond{60.1} & \fprsecond{42.0}
& 76.9 & 72.2
& \fprsecond{70.3} & \fprsecond{61.3}
& \fprsecond{65.5} & \fprsecond{52.2}
& \fprsecond{65.9} & \fprsecond{53.1} \\

\midrule

\multicolumn{23}{@{}l@{}}{\textbf{10\% FPR}\quad\textit{Supervised detectors}} \\
\midrule

RADAR
& 50.4 & 16.5
& 47.8 & 8.8
& 48.7 & 11.6
& 46.5 & 4.8
& 47.1 & 7.5
& 47.5 & 8.1
& 47.8 & 9.2
& 52.1 & 18.9
& 46.6 & 5.3
& 45.9 & 3.2
& 45.9 & 3.3 \\

ReMoDetect
& \fprsecond{74.4} & \fprsecond{62.9}
& \fprbest{62.6} & \fprsecond{42.9}
& \fprbest{60.7} & \fprbest{39.5}
& \fprbest{70.8} & \fprbest{57.2}
& \fprbest{67.0} & \fprbest{51.2}
& \fprbest{63.4} & \fprbest{44.8}
& \fprbest{68.2} & \fprbest{53.5}
& 75.4 & 64.9
& \fprbest{72.4} & \fprbest{57.1}
& \fprbest{71.2} & \fprbest{53.7}
& \fprbest{71.0} & \fprbest{53.7} \\

Longformer
& 71.5 & 62.6
& 50.6 & 17.7
& 49.9 & 15.5
& 51.2 & 19.7
& 50.1 & 16.8
& 49.8 & 15.3
& 50.8 & 18.3
& \fprsecond{76.2} & \fprsecond{70.7}
& 58.2 & 35.8
& 56.3 & 31.5
& 56.6 & 31.9 \\

BLOOMZ-3B
& \fprbest{89.7} & \fprbest{88.6}
& \fprsecond{61.9} & \fprbest{43.5}
& \fprsecond{58.3} & \fprsecond{35.6}
& \fprsecond{62.4} & \fprsecond{44.5}
& \fprsecond{59.9} & \fprsecond{38.9}
& 52.8 & \fprsecond{22.7}
& 51.8 & \fprsecond{21.4}
& \fprbest{94.3} & \fprbest{95.0}
& \fprsecond{61.6} & \fprsecond{44.1}
& \fprsecond{59.8} & \fprsecond{39.5}
& \fprsecond{60.2} & \fprsecond{40.0} \\

OpenAI RoBERTa
& 48.1 & 10.1
& 48.4 & 10.0
& 48.7 & 10.9
& 47.8 & 8.4
& 47.0 & 6.8
& 50.8 & 16.6
& 51.5 & 18.4
& 48.2 & 10.0
& 47.3 & 7.7
& 47.5 & 7.9
& 47.7 & 8.4 \\

ConDA
& 49.8 & 14.9
& 49.5 & 14.4
& 49.4 & 13.9
& 49.8 & 15.5
& 48.5 & 11.6
& 47.8 & 9.1
& 47.6 & 8.8
& 51.0 & 18.0
& 53.8 & 25.4
& 51.2 & 19.2
& 50.9 & 18.2 \\

ChatGPT Detector
& 55.5 & 26.7
& 52.6 & 19.1
& 51.0 & 15.6
& 56.7 & 29.0
& 53.7 & 22.7
& \fprsecond{52.9} & 19.9
& \fprsecond{53.2} & 21.2
& 55.3 & 26.2
& 56.7 & 29.1
& 56.9 & 27.3
& 57.0 & 27.2 \\

\bottomrule
\end{tabular*}
\endgroup
\end{table}

\noindent\textbf{A strict false-positive budget makes collaborative text selectively invisible.}
Table~\ref{tab:finegrained_acc_f1_all_fpr} shows that at 1\% FPR, the strongest detectors retain relatively high ACC and F1 on MGT and L1, whereas content filling, machine rewriting, and more complex prompt conditions generally produce lower scores. The degradation is more pronounced in F1 than in ACC, indicating that the strict operating point suppresses true-positive coverage even when the overall accuracy remains close to chance. This pattern suggests that shallow collaboration can preserve detectable machine regularities, while deeper human intervention or stronger generation constraints weaken the evidence available to the detector.

\noindent\textbf{Increasing the FPR buys detection coverage, but not recovered evidence.}
Relaxing the FPR from 1\% to 5\% and 10\% generally improves both ACC and F1, particularly for collaborative conditions that perform poorly under the strictest threshold. However, the relative difficulty across collaboration conditions remains uneven, and the strongest performance does not converge to a common ordering. Increasing the FPR changes the decision threshold and allows weaker machine-generation scores to be accepted, but it cannot restore signals that collaboration has removed or redistributed. Thus, a more permissive operating point trades false positives for coverage rather than resolving the underlying distribution shift.

\noindent\textbf{Supervision redistributes detector sensitivity rather than removing distribution shift.}
The upper-performing zero-shot detectors maintain a relatively coherent performance pattern across FPR levels, whereas supervised detectors exhibit condition-dependent crossovers. BLOOMZ-3B performs particularly well on MGT and L1, but loses this advantage on many collaborative conditions, where ReMoDetect becomes more competitive. This contrast indicates that supervision does not produce a collaboration-invariant decision boundary; instead, it transfers the training distribution's preferences into the detector. The decisive factor is therefore not whether a detector is supervised, but whether its learned or intrinsic evidence remains aligned with the collaboration process under evaluation.

\subsection{Supplementary Analysis of Domain Sensitivity}
\label{app:domain_results}

\begin{table}[H]
\centering
\caption{Domain-wise AUROC of zero-shot and off-the-shelf supervised detectors. Each value is macro-averaged over six generation models and the fine-grained conditions within each of the four text-production settings, followed by an equal-weight average over the four settings. Within each detector family and domain, the best and second-best results are shown in boldface and underlined, respectively.}
\begingroup
\fontsize{8.5}{10}\selectfont
\setlength{\tabcolsep}{4pt}
\renewcommand{\arraystretch}{1.04}
\begin{tabularx}{\linewidth}{@{}l*{5}{>{\centering\arraybackslash}X}@{}}
\toprule
\textbf{Detector} & \textbf{Law} & \textbf{Literature} & \textbf{News} & \textbf{Encyclopedia} & \textbf{Q\&A} \\
\midrule
\multicolumn{6}{@{}l@{}}{\textit{Zero-shot detectors}} \\
\midrule
Log-Likelihood & 0.891 & 0.575 & 0.618 & 0.676 & 0.698 \\
LogRank & 0.873 & 0.583 & 0.614 & 0.669 & 0.698 \\
Entropy & 0.182 & 0.479 & 0.399 & 0.346 & 0.337 \\
DetectLLM-LRR & 0.663 & 0.602 & 0.572 & 0.602 & 0.661 \\
Fast-DetectGPT & 0.783 & 0.602 & 0.552 & 0.604 & 0.655 \\
Binoculars & 0.801 & \textbf{0.809} & 0.662 & 0.736 & 0.702 \\
LastDE & 0.832 & \underline{0.686} & 0.766 & 0.768 & 0.749 \\
LastDE++ & 0.876 & 0.685 & 0.728 & 0.761 & 0.788 \\
SpecDetect & \textbf{0.958} & 0.675 & \underline{0.790} & \textbf{0.875} & \textbf{0.868} \\
SpecDetect++ & \underline{0.905} & 0.682 & \textbf{0.794} & \underline{0.841} & \underline{0.836} \\
\midrule
\multicolumn{6}{@{}l@{}}{\textit{Supervised detectors}} \\
\midrule
RADAR & 0.529 & 0.580 & 0.397 & 0.421 & 0.268 \\
ReMoDetect & \textbf{0.923} & 0.627 & \underline{0.870} & \underline{0.854} & 0.690 \\
Longformer & 0.663 & \textbf{0.698} & 0.686 & 0.553 & 0.667 \\
BLOOMZ-3B & \underline{0.837} & 0.594 & \textbf{0.894} & \textbf{0.874} & \textbf{0.858} \\
OpenAI RoBERTa & 0.508 & 0.572 & 0.502 & 0.390 & 0.441 \\
ConDA & 0.647 & 0.572 & 0.439 & 0.450 & 0.470 \\
ChatGPT Detector & 0.759 & \underline{0.653} & 0.672 & 0.621 & \underline{0.719} \\
\bottomrule
\end{tabularx}
\endgroup

\label{tab:domain_auroc_combined}
\end{table}

\noindent\textbf{Domain variation creates a structured but uneven detection landscape.} Table~\ref{tab:domain_auroc_combined} summarizes domain-wise AUROC after averaging over generation models and fine-grained collaboration conditions. Law is generally the most detectable domain, whereas Literature and News tend to yield lower scores. This pattern indicates that domain characteristics systematically affect the separability between human-written and machine-generated text: the constrained vocabulary and discourse structure of legal text may expose generation artifacts more consistently, while the broader stylistic variation of open-ended domains increases the overlap between the two classes.

\noindent\textbf{Zero-shot detectors exhibit relatively stable domain sensitivity because they rely on shared statistical evidence.} Their domain ordering remains broadly consistent across text-production settings, with SpecDetect and SpecDetect++ achieving strong overall performance and Binoculars showing a notable advantage on Literature. Although their scoring functions differ, these detectors primarily exploit probability distributions, token ranks, or related distributional statistics derived from language models. Domain-specific regularities therefore modulate a common type of evidence: constrained legal writing can amplify deviations in model-based statistics, whereas more stylistically diverse domains weaken their consistency.

\noindent\textbf{Supervised detectors transform domain shifts into model-specific weaknesses.} The strongest detector changes across domains: ReMoDetect performs best on Law, BLOOMZ-3B dominates News, Encyclopedia, and Q\&A, while Longformer is strongest on Literature. This variation reflects the dependence of supervised decision boundaries on training data, model architecture, and optimization objectives. Consequently, detector reliability cannot be inferred from an aggregate AUROC alone; it depends on whether the learned cues remain aligned with the target domain.

\begin{table}[H]
\centering
\caption{Detailed boundary-localization performance of the two fine-tuned detectors on the common content-filling test set. Here, $n$ denotes the number of evaluated documents, and values are document-level means $\pm$ population standard deviations. Boundary F1@1 uses one-to-one matching between predicted and gold boundaries within a tolerance of one sentence; arrows indicate the preferred direction.}
\label{tab:boundary_localization_detailed}
\scriptsize
\setlength{\tabcolsep}{2.5pt}
\renewcommand{\arraystretch}{1.08}
\begin{tabular}{@{}llcccc@{}}
\toprule
\textbf{Condition} & \textbf{Detector} & \textbf{$n$} & \textbf{Sentence $\kappa$ $\uparrow$} & \textbf{WindowDiff $\downarrow$} & \textbf{Boundary F1@1 $\uparrow$} \\
\midrule
LC-C & WCP & 1962 & 0.194 $\pm$ 0.343 & 0.428 $\pm$ 0.206 & 0.168 $\pm$ 0.226 \\
LC-C & DeBERTa--BiGRU--CRF & 1962 & -0.054 $\pm$ 0.101 & 0.386 $\pm$ 0.184 & 0.004 $\pm$ 0.051 \\
\midrule
LC-D & WCP & 1886 & 0.019 $\pm$ 0.163 & 0.381 $\pm$ 0.101 & 0.380 $\pm$ 0.232 \\
LC-D & DeBERTa--BiGRU--CRF & 1886 & 0.258 $\pm$ 0.250 & 0.309 $\pm$ 0.102 & 0.373 $\pm$ 0.271 \\
\midrule
HC-C & WCP & 1877 & 0.166 $\pm$ 0.373 & 0.451 $\pm$ 0.185 & 0.171 $\pm$ 0.231 \\
HC-C & DeBERTa--BiGRU--CRF & 1877 & 0.005 $\pm$ 0.152 & 0.336 $\pm$ 0.089 & 0.341 $\pm$ 0.327 \\
\midrule
HC-D & WCP & 1637 & 0.051 $\pm$ 0.198 & 0.355 $\pm$ 0.099 & 0.412 $\pm$ 0.238 \\
HC-D & DeBERTa--BiGRU--CRF & 1637 & 0.238 $\pm$ 0.183 & 0.303 $\pm$ 0.101 & 0.464 $\pm$ 0.232 \\
\midrule
Overall & WCP & 7362 & 0.110 $\pm$ 0.297 & 0.405 $\pm$ 0.162 & 0.277 $\pm$ 0.258 \\
Overall & DeBERTa--BiGRU--CRF & 7362 & 0.106 $\pm$ 0.227 & 0.335 $\pm$ 0.131 & 0.287 $\pm$ 0.299 \\
\bottomrule
\end{tabular}
\end{table}
\begin{table}[H]
\centering
\caption{Boundary-prediction diagnostics for the two fine-tuned detectors on the common content-filling test set. Here, $n$ denotes the number of evaluated documents. No predicted boundary is the percentage of documents with zero predicted boundaries; Mean predicted and Mean gold are the document-level average numbers of predicted and gold boundaries, respectively. Mean count error is defined as gold minus predicted boundaries, so negative and positive values indicate over-segmentation and missed boundaries, respectively.}
\label{tab:boundary_prediction_diagnostics}
\scriptsize
\setlength{\tabcolsep}{2.5pt}
\renewcommand{\arraystretch}{1.08}
\begin{tabular}{@{}llccccc@{}}
\toprule
\textbf{Condition} & \textbf{Detector} & \textbf{$n$} & \textbf{No predicted boundary (\%)} & \textbf{Mean predicted} & \textbf{Mean gold} & \textbf{Mean count error} \\
\midrule
LC-C & WCP & 1962 & 4.18 & 4.497 & 1.000 & -3.497 \\
LC-C & DeBERTa--BiGRU--CRF & 1962 & 70.90 & 0.629 & 1.000 & 0.371 \\
\midrule
LC-D & WCP & 1886 & 5.51 & 4.073 & 4.914 & 0.841 \\
LC-D & DeBERTa--BiGRU--CRF & 1886 & 24.23 & 1.694 & 4.914 & 3.221 \\
\midrule
HC-C & WCP & 1877 & 3.94 & 4.384 & 1.000 & -3.384 \\
HC-C & DeBERTa--BiGRU--CRF & 1877 & 31.43 & 1.477 & 1.000 & -0.477 \\
\midrule
HC-D & WCP & 1637 & 6.66 & 3.664 & 5.561 & 1.897 \\
HC-D & DeBERTa--BiGRU--CRF & 1637 & 2.69 & 2.561 & 5.561 & 2.999 \\
\midrule
Overall & WCP & 7362 & 5.01 & 4.174 & 3.017 & -1.158 \\
Overall & DeBERTa--BiGRU--CRF & 7362 & 33.71 & 1.548 & 3.017 & 1.469 \\
\bottomrule
\end{tabular}
\end{table}

\subsection{Supplementary Results for Boundary Localization}
\label{app:boundary-localization}

\noindent\textbf{Comparable overall scores conceal opposite boundary-count biases.}
As shown in Table~\ref{tab:boundary_localization_detailed}, the two detectors achieve nearly identical overall Sentence $\kappa$ values, while DeBERTa--BiGRU--CRF obtains a lower WindowDiff and a slightly higher Boundary F1@1. However, the diagnostics in Table~\ref{tab:boundary_prediction_diagnostics} reveal substantially different boundary-count errors: WCP tends to over-segment, whereas DeBERTa--BiGRU--CRF predicts fewer boundaries and produces no predicted boundary for 33.71\% of documents. Thus, similar aggregate scores correspond to different localization behaviors.

\noindent\textbf{Collaboration structure changes the dominant localization error.}
Table~\ref{tab:boundary_prediction_diagnostics} shows that the C conditions contain approximately one gold boundary per document. Under these conditions, WCP produces more than four predicted boundaries per document, indicating strong over-segmentation. In the denser D conditions, its mean count error becomes positive, suggesting that the detector does not fully recover the increased number of author transitions. DeBERTa--BiGRU--CRF exhibits a more conservative pattern: it predicts no boundary for 70.90\% of LC-C documents, but this proportion decreases to 2.69\% under HC-D. Nevertheless, its predicted boundary count remains below the gold count in the D conditions.

\noindent\textbf{Boundary localization requires joint interpretation of quality and count consistency.}
The results in Tables~\ref{tab:boundary_localization_detailed} and~\ref{tab:boundary_prediction_diagnostics} show why a single metric is insufficient. DeBERTa--BiGRU--CRF achives a lower overall WindowDiff and a slightly higher Boundary F1@1 in Table~\ref{tab:boundary_localization_detailed}, but its positive overall count error in Table~\ref{tab:boundary_prediction_diagnostics} indicates systematic under-segmentation. Conversely, WCP's higher predicted boundary count reveals over-segmentation that is not fully reflected by its aggregate F1 score. These results indicate that boundary localization should be evaluated jointly in terms of boundary matching and preservation of the overall author-transition structure.

\subsection{Supplementary Cross-Condition Generalization Results}
\label{app:cross-detection}

\begin{figure}[H]
    \centering
    \includegraphics[width=0.75\linewidth]{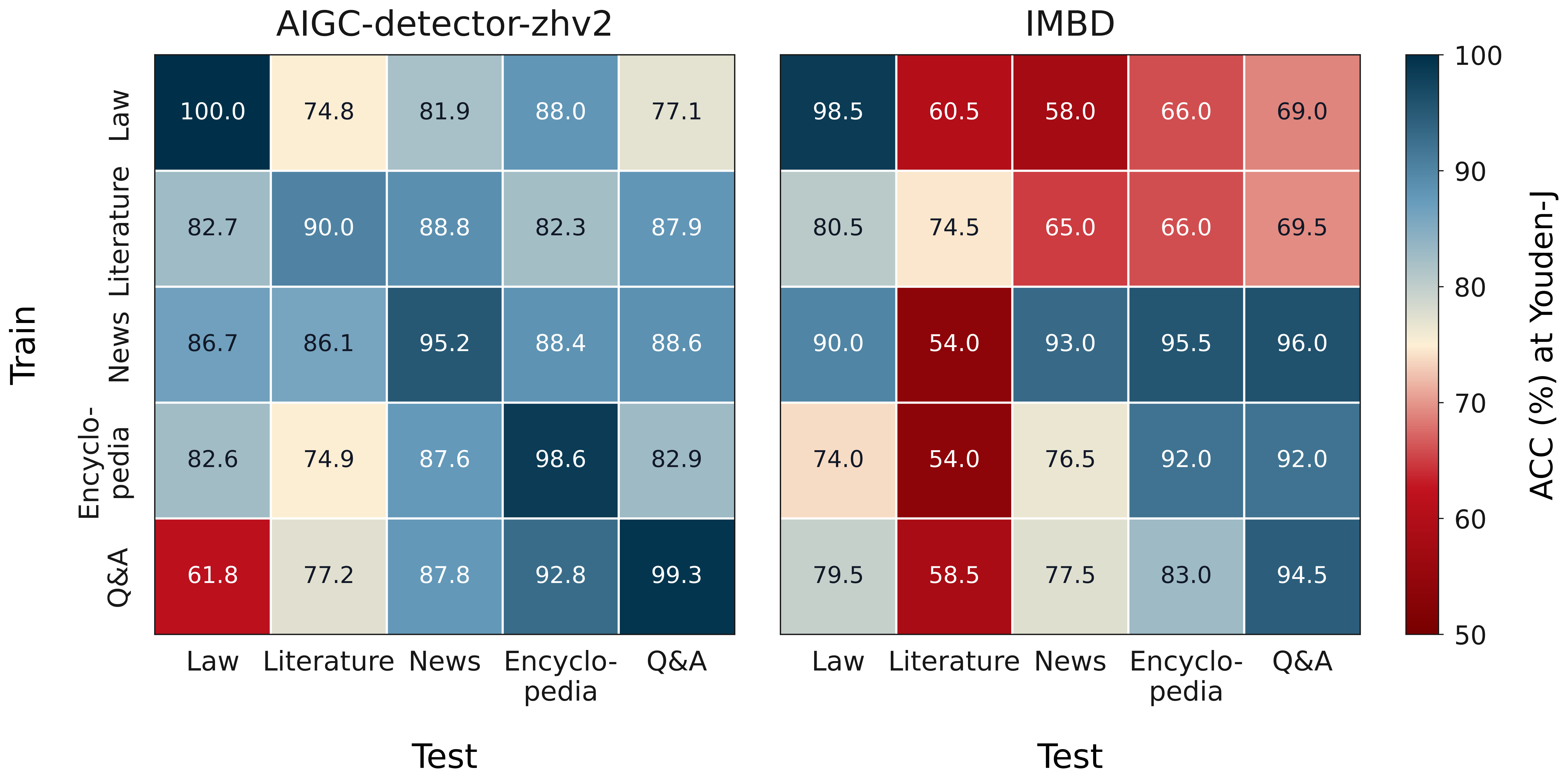}
    \caption{Cross-domain accuracy (\%) of AIGC-detector-zhv2 and IMBD with cellwise Youden-optimal thresholds. Rows and columns denote training and test domains, respectively.}
    \label{fig:cross-domain}
\end{figure}

\begin{figure}[H]
    \centering
    \includegraphics[width=0.7\linewidth]{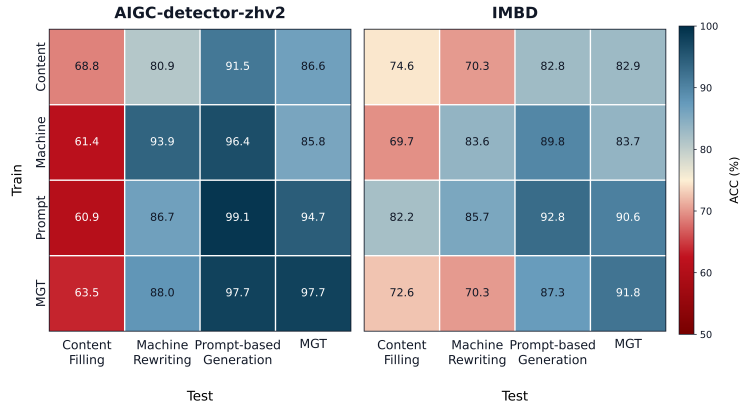}
    \caption{Accuracy (\%) across collaboration types for AIGC-detector-zhv2 and IMBD with cellwise Youden-optimal thresholds. Rows and columns denote training and test collaboration types, respectively.}
    \label{fig:cross-methods}
\end{figure}

\noindent\textbf{Domain transfer is asymmetric across source and target domains.}
Figure~\ref{fig:cross-domain} shows that News provides the strongest cross-domain training source for both detectors, whereas Literature is the most difficult target, especially for IMBD. This source--target asymmetry suggests that transfer depends on the compatibility between the cues learned from the source domain and the distribution of the target domain, rather than on in-domain accuracy alone.

\noindent\textbf{Content Filling is a detector-specific transfer bottleneck.}
In Figure~\ref{fig:cross-methods}, Content Filling is the weakest target for AIGC-detector-zhv2, while IMBD retains stronger transfer to this target when trained on Prompt-based Generation. The contrast indicates that the difficulty of detecting locally inserted machine content is not determined solely by the production mode, but also by the cues learned by each detector.

\begin{figure}[H]
    \centering
    \includegraphics[width=0.75\linewidth]{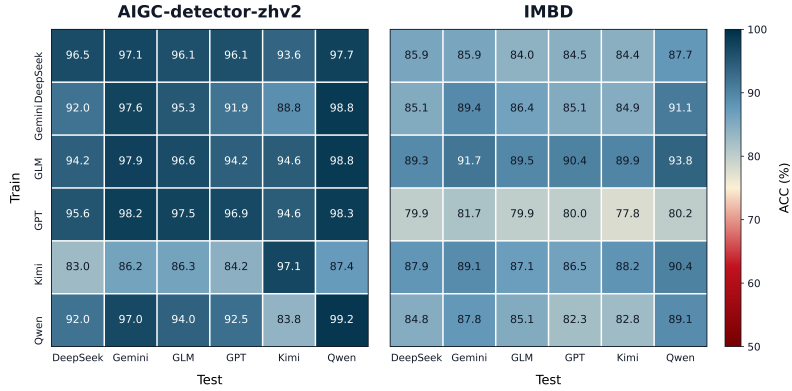}
    \caption{Cross-generator accuracy (\%) of AIGC-detector-zhv2 and IMBD with cellwise Youden-optimal thresholds. Rows and columns denote training and test generators, respectively.}
    \label{fig:cross-model}
\end{figure}

\noindent\textbf{Transfer utility and target difficulty are not equivalent.}
For AIGC-detector-zhv2, training on Content Filling transfers well to Prompt-based Generation and MGT, whereas the reverse directions perform poorly on Content Filling. IMBD exhibits a weaker version of this asymmetry. Thus, a condition may be difficult to recognize as an unseen target while still providing useful training signals for other conditions, making train--test mode matching insufficient for predicting transfer quality.

\noindent\textbf{Cross-generator transfer is governed by detector--source interactions.}
Figure~\ref{fig:cross-model} shows high in-generator performance for both detectors, but off-diagonal transfer varies substantially with the training generator. GPT is the strongest source for AIGC-detector-zhv2 but the weakest source for IMBD, while Kimi causes a marked source-side drop for AIGC-detector-zhv2 without a comparable effect on IMBD. These results show that matched generators do not guarantee the best transfer, and that the transfer value of a generator cannot be assessed independently of the detector.

\section{Representative Data Construction Examples}
\label{app:data-examples}
\begin{CJK*}{UTF8}{gbsn}
\begin{center}
\small
\begin{minipage}{1.00\linewidth}
We present one human-written news article and its eleven paired variants from C-HAT-240K, all sharing \textit{source\_id} \textit{0001}. The variants cover MGT, four Content Filling conditions, two Machine Rewriting levels, and four Prompt-based Generation levels.
\end{minipage}
\end{center}
\medskip
\begin{promptbox}{Paired Data Example from the News Domain}
\textbf{Human-written Source}\par
\textit{Original human text}\par\smallskip
帕切科用对讲机指挥球队被禁 传达战术只能靠纸条 新浪体育讯 当帕切科被停赛8场之后，所有人最关心的一个问题就是帕切科到底如何指挥国安，尤其是在不能依靠对讲机、手机等现代通讯的情况下，帕切科如何指挥成了一个大问题。就在7月2日与陕西的比赛之中，更让人意料不到的是，帕切科竟然在比赛开始之前“消失”了，那么帕切科到底是如何指挥队员们完成这场最艰苦的比赛呢？比赛之前的联席会上，国安队领队魏克兴被明确告之，替补席上的教练员不得使用对讲机与外界进行沟通，因为这个规定是之前早就做出的规定，并不是针对帕切科的停赛，所有球队的替补席人员都不能使用对讲机。这样一来，就比较麻烦了，国安带到西安的对讲机也用不上。虽然足协没有对使用手机做出限制，但是考虑到比赛现场的信号不好，而且现场特别吵杂，手机通话效果无法保证，最终教练组决定用两个办法传递帕切科的指挥：第一，帕切科把自己的想法写在小纸条，然后传递给教练组；还有就是通过发手机短信的形式发到路易斯的手机上，但是考虑到怕信号不好，手机短信会有延迟，还是决定主要以小纸条这种“鸡毛信”的方式为第一方案，发短信为第二方案。确定了这两个方案之后，帕切科坐在哪里又成为一个大问题。如果是在工体，帕切科可以坐在主席台或者贵宾席上，但在客场，尤其是是西安这个客场，帕切科如果坐在看台上、甚至主席台上都有可能会受到西安球迷的干扰。为了能够让帕切科有一个安静的环境，国安与陕西方面进行了反复的沟通，最终把帕切科安排在了体育场最高处平时几个用来转播的小房间其中之一，而这个小房间也是现场转播的死角，因此在这场比赛之中，不仅现场看不到帕切科坐在哪里，电视镜头也无法捕捉到帕切科的身影，老帕就这样神秘“消失”了。在比赛之中，国安专门准备了两个工作人员传递鸡毛信，其中一个人陪在帕切科身边，在帕切科写完小纸条之后，立刻做电梯来到休息室，再由休息室外的副领队康玉明把小纸条传递给替补席的路易斯。除此之外，帕切科分别对场每一个位置的对应换人进行了交待，这也是在张辛昕受伤情况不明之时，替补席立刻招来小将姜涛准备替补的原因。比赛结束之后，帕切科也承认，自己不能在替补席指挥比赛肯定会给球队造成一定影响，但这种经历自己过去也有过，自己相信任何一个职业教练在执教生涯之中都会遇到这种现象，所以自己也会有相应的办法应对，今天路易斯和其他中方教练代替自己在替补席完成了指挥工作，自己很感谢他们。(宋词 发自西安)
\par\medskip\hrule\medskip
\textbf{MGT}\par
\textit{Prefix continuation}\par\smallskip
帕切科用对讲机指挥球队被禁 传达战术只能靠纸条 新浪体育讯 当帕切科被停赛之后，国安教练组如何在比赛中完成临场指挥，便成了外界关注的焦点。按照足协纪律规定，停赛教练不得进入替补席和更衣室，也不能通过电子通讯设备与场内人员保持联系。也就是说，此前被一些球队采用的对讲机、手机短信等方式都被明确禁止，帕切科即便坐在看台上，也只能像普通观众一样观看比赛。在昨天下午的训练中，国安方面已经开始演练“无主帅指挥”的应急方案。助理教练将在替补席上负责临场调度，而帕切科赛前准备的战术意图，则会通过书面材料提前交给教练组。比赛过程中如果需要调整，教练席只能根据赛前预案和场上形势自行判断。俱乐部人士表示：“规则怎么要求，我们就怎么执行，肯定不会冒险违规。”不过，帕切科一向以激情和临场调整见长，不能在场边大声指挥，对国安显然会有影响。尤其是球队在比赛僵持阶段，换人时机、阵型变化都需要果断决定。为了尽量减少损失，教练组这两天反复研究对手录像，并把可能出现的几种局面列成方案，细化到领先、落后、少打一人等不同情况。队员们也被要求在场上提高自主判断能力，几名老队员将承担更多提醒和组织责任。按照安排，帕切科赛前仍可以参加最后一次战术会议，但在比赛开始前必须离开限制区域。届时他很可能出现在看台包厢内观看比赛。由于不能使用对讲机传达指令，如果确有需要，工作人员也只能在不违反规定的前提下，通过中场休息时的书面意见进行沟通。有人开玩笑说，国安这次恐怕要重新回到“纸条时代”。事实上，足协近来对停赛教练的管理比以往更加严格，目的就是防止处罚流于形式。此前有教练被罚上看台后仍通过耳机指挥，曾引发争议。相关人士强调，停赛的意义就在于剥夺其现场执教权，如果仍能实时控制替补席，处罚就失去了效果。因此本轮比赛监督也会特别留意国安替补席的通讯情况。对于即将到来的比赛，国安队内的态度比较平静。一名队员表示，主教练不能站在场边当然会不习惯，但球队这一周训练安排得很细，大家清楚自己该怎么踢。“帕切科的要求已经讲了很多遍，关键还是我们在场上执行。”俱乐部也希望借此机会证明，球队并不只依赖主帅的临场吼叫，而是有一套相对成熟的战术体系。帕切科本人则没有公开抱怨处罚。熟悉他的人透露，葡萄牙人最担心的不是自己不能指挥，而是球员因此分心。他在训练后特意叮嘱队员，把注意力放在比赛本身，不要被场外话题影响。至于那些写满战术要点的纸条，能否在关键时刻发挥作用，只能等比赛开始后由结果来回答。
\par\medskip\hrule\medskip
\textbf{Content Filling}\par
\textit{LC\_D, dispersed filling}\par\smallskip
帕切科用对讲机指挥球队被禁 传达战术只能靠纸条 新浪体育讯 当帕切科被停赛8场之后，所有人最关心的一个问题就是帕切科到底如何指挥国安，尤其是在不能依靠对讲机、手机等现代通讯的情况下，帕切科如何指挥成了一个大问题。就在7月2日与陕西的比赛之中，更让人意料不到的是，帕切科竟然在比赛开始之前“消失”了，那么帕切科到底是如何指挥队员们完成这场最艰苦的比赛呢？比赛之前的联席会上，国安队领队魏克兴被明确告之，替补席上的教练员不得使用对讲机与外界进行沟通，这也就意味着帕切科即便不坐在替补席上，也不能像以往那样通过对讲机把指令直接传到场边。至于能否打手机，赛前并没有得到明确许可，教练组也不敢冒险，而且现场特别吵杂，手机通话效果无法保证，最终教练组决定用两个办法传递帕切科的指挥：第一，帕切科把自己的想法写在小纸条，然后传递给教练组；还有就是通过发手机短信的形式发到路易斯的手机上，但是考虑到怕信号不好，手机短信会有延迟，还是决定主要以小纸条这种“鸡毛信”的方式为第一方案，发短信为第二方案。确定了这两个方案之后，帕切科坐在哪里又成为一个大问题。如果是在工体，帕切科可以坐在主席台或者贵宾席上，但在客场，尤其是是西安这个客场，帕切科如果坐在看台上、甚至主席台上都有可能会受到西安球迷的干扰。为了能够让帕切科有一个安静的环境，国安与陕西方面进行了反复的沟通，最终把帕切科安排在了体育场最高处平时几个用来转播的小房间其中之一，而这个小房间也是现场转播的死角，因此在这场比赛之中，不仅现场看不到帕切科坐在哪里，电视镜头也无法捕捉到帕切科的身影，不过在这个位置上，帕切科可以俯瞰全场，国安也专门安排工作人员在房间与替补席之间往返，他把需要提醒的站位、换人和攻防细节写在纸上后，马上由工作人员送到场边，再由路易斯和中方教练向队员传达。由于视野开阔，场上每一个细节都能被他及时发现，这也是在张辛昕受伤情况不明之时，替补席立刻招来小将姜涛准备替补的原因。比赛结束之后，帕切科也承认，自己不能在替补席指挥比赛肯定会给球队造成一定影响，但这种经历自己过去也有过，自己相信任何一个职业教练在执教生涯之中都会遇到这种现象，所以自己也会有相应的办法应对，今天路易斯和其他中方教练代替自己在替补席完成了指挥工作，自己很感谢他们。(宋词 发自西安)
\par\medskip
\textit{LC\_C, continuous filling}\par\smallskip
帕切科用对讲机指挥球队被禁 传达战术只能靠纸条 新浪体育讯 当帕切科被停赛8场之后，所有人最关心的一个问题就是帕切科到底如何指挥国安，尤其是在不能依靠对讲机、手机等现代通讯的情况下，帕切科如何指挥成了一个大问题。就在7月2日与陕西的比赛之中，更让人意料不到的是，帕切科竟然在比赛开始之前“消失”了，那么帕切科到底是如何指挥队员们完成这场最艰苦的比赛呢？比赛之前的联席会上，国安队领队魏克兴被明确告之，替补席上的教练员不得使用对讲机与外界进行沟通，因为这个规定是之前早就做出的规定，并不是针对帕切科的停赛，所有球队的替补席人员都不能使用对讲机。这样一来，就比较麻烦了，国安带到西安的对讲机也用不上。虽然足协没有对使用手机做出限制，但是考虑到比赛现场的信号不好，而且现场特别吵杂，手机通话效果无法保证，最终教练组决定用两个办法传递帕切科的指挥：第一，帕切科把自己的想法写在小纸条，然后传递给教练组；还有就是通过发手机短信的形式发到路易斯的手机上，但是考虑到怕信号不好，手机短信会有延迟，还是决定主要以小纸条这种“鸡毛信”的方式为第一方案，发短信为第二方案。确定了这两个方案之后，帕切科坐在哪里又成为一个大问题。如果是在工体，帕切科可以坐在主席台或者贵宾席上，但在客场，尤其是是西安这个客场，帕切科如果坐在看台上、甚至主席台上都有可能会受到西安球迷的干扰。为了能够让帕切科有一个安静的环境，国安与陕西方面进行了反复的沟通，最终把帕切科安排在了体育场最高处平时几个用来转播的小房间其中之一，而这个小房间也是现场转播的死角，因此在这场比赛之中，不仅现场看不到帕切科坐在哪里，电视镜头也无法捕捉到帕切科的身影，老帕就这样神秘“消失”了。比赛开始之后，帕切科就在这个小房间里通过电视转播和现场视角观察比赛，一旦有新的要求，便立即写在纸条上，由身边的工作人员送到看台通道，再交给守候在附近的国安工作人员，最后传到替补席的教练组手中。由于小房间距离替补席并不算近，每一次传递都需要一定时间，所以帕切科的指令也不可能像平时那样随时、迅速地传达。相比之下，中场休息时的沟通就显得更为重要，教练组把上半场出现的问题集中反馈给他，帕切科再把下半场的调整意见交代清楚。就这样，国安在没有主教练现场坐镇、也无法使用对讲机的情况下，只能依靠这种最原始的办法完成临场指挥。
\par\medskip
\textit{HC\_D, dispersed filling}\par\smallskip
帕切科用对讲机指挥球队被禁 传达战术只能靠纸条 新浪体育讯 当帕切科被停赛8场之后，所有人最关心的一个问题就是帕切科到底如何指挥国安，国安教练组最初想到的办法，是让老帕坐在看台上通过对讲机随时把战术意图传到替补席，这也是很多人认为最直接、最有效的方式。为了确认这一做法是否可行，俱乐部赛前也专门向比赛监督进行了询问，但得到的答复是否定的：停赛教练不能以任何形式在场边进行指挥，同时按照中超联赛有关规定，所有球队的替补席人员都不能使用对讲机。这样一来，就比较麻烦了，国安带到西安的对讲机也用不上。虽然足协没有对使用手机做出限制，但是考虑到比赛现场的信号不好，而且现场特别吵杂，手机通话效果无法保证，最终教练组决定用两个办法传递帕切科的指挥：第一，帕切科把自己的想法写在小纸条，然后传递给教练组；还有就是通过发手机短信的形式发到路易斯的手机上，但是考虑到怕信号不好，所以最可靠的办法还是“纸条传令”。不过，用纸条也必须解决一个问题：帕切科到底坐在什么位置，既能够看清场上情况，又能保证纸条尽快送到替补席。赛前，国安方面为此费了不少周折，为了能够让帕切科有一个安静的环境，国安与陕西方面进行了反复的沟通，最终把帕切科安排在了体育场最高处平时几个用来转播的小房间其中之一，而这个小房间也是现场转播的死角，因此在这场比赛之中，不仅现场看不到帕切科坐在哪里，电视镜头也无法捕捉到帕切科的身影，老帕就这样神秘“消失”了。在比赛之中，国安专门准备了两个工作人员传递鸡毛信，其中一个人陪在帕切科身边，在帕切科写完小纸条之后，立刻做电梯来到休息室，再把纸条交给等候在那里的另一名工作人员，由他迅速送到替补席交给路易斯。由于从看台最高处到替补席路程不短，这种传递方式难免有些滞后，工作人员还要避开球迷和现场安保通道，因此老帕在上半时很少频繁调整，更多是在中场休息时把自己的意见集中告诉教练组。不过从比赛过程看，替补席对他的意图执行得比较坚决。赛后谈到这种特殊的指挥方式，帕切科表示，今天路易斯和其他中方教练代替自己在替补席完成了指挥工作，自己很感谢他们。(宋词 发自西安)
\par\medskip
\textit{HC\_C, continuous filling}\par\smallskip
帕切科用对讲机指挥球队被禁 传达战术只能靠纸条 新浪体育讯 当帕切科被停赛8场之后，所有人最关心的一个问题就是帕切科到底如何指挥国安，尤其是在不能依靠对讲机、手机等现代通讯的情况下，帕切科如何指挥成了一个大问题。就在7月2日与陕西的比赛之中，更让人意料不到的是，帕切科竟然在比赛开始之前“消失”了，那么帕切科到底是如何指挥队员们完成这场最艰苦的比赛呢？比赛之前的联席会上，国安队领队魏克兴被明确告之，替补席上的教练员不得使用对讲机与外界进行沟通，因为这个规定是之前早就做出的规定，并不是针对帕切科的停赛，所有球队的替补席人员都不能使用对讲机。这样一来，就比较麻烦了，国安带到西安的对讲机也用不上。按照足协的处罚规定，帕切科不能进入替补席，也不能在赛前、赛中进入休息室进行布置，因此国安方面只能提前把各种可能出现的情况都做成预案。赛前踩场和准备会，助理教练已经把首发、换人以及定位球安排传达给队员，真正到了比赛中，如果临场出现变化，就只能靠看台上的观察和替补席之间最原始的方式联系。开赛前，记者在贵宾席和看台上都没有看到帕切科的身影，甚至连国安工作人员也一度表示“不知道他在哪里”。原来，为了避免引起不必要的争议，帕切科并没有像外界想象的那样坐在醒目的位置遥控指挥，而是选择在一个相对隐蔽的地方观看比赛。由于不能使用对讲机，他只能把自己的意见写在纸上，再由工作人员辗转送到替补席。这样的传递显然无法做到及时，往往等纸条送到场边时，场上的形势已经发生了变化。比赛中，替补席上的助理教练更多是按照赛前既定部署来调整。上半场国安被对手压制时，帕切科显得非常着急，但他能做的也只是把对防守站位和中场逼抢的提醒写下来，让人送下去。到了下半场，国安需要换人加强进攻，相关意见同样是通过纸条完成。相比过去帕切科在场边大声吼叫、不断用手势提醒队员，这种方式无疑让他的作用被大大削弱。不过，国安队员赛后表示，虽然主教练不在身边，但大家赛前已经清楚他的要求，“他平时训练里强调得很多，所以比赛中还是按照他的思路去踢。”对于接下来多场比赛仍将面临同样局面，国安俱乐部也只能继续寻找合规的办法，在不违反纪律的前提下尽量保证教练组与球队之间的沟通。
\par\medskip\hrule\medskip
\textbf{Machine Rewriting}\par
\textit{Light Rewriting}\par\smallskip
帕切科用对讲机指挥球队被禁 传达战术只能靠纸条 新浪体育讯 帕切科被停赛8场之后，所有人最关心的问题就是他将如何指挥国安。尤其是在不能依靠对讲机、手机等现代通讯工具的情况下，帕切科如何进行临场指挥成了一个大问题。7月2日与陕西的比赛中，更让人意想不到的是，帕切科竟然在比赛开始前“消失”了。那么，他到底是如何指挥队员们完成这场艰苦比赛的呢？比赛之前的联席会上，国安队领队魏克兴被明确告知，替补席上的教练员不得使用对讲机与外界沟通。因为这一规定早已做出，并不是针对帕切科的停赛，所有球队的替补席人员都不能使用对讲机。这样一来，事情就变得比较麻烦，国安带到西安的对讲机也用不上了。虽然足协没有对使用手机做出限制，但考虑到比赛现场信号不好，而且现场特别嘈杂，手机通话效果无法保证，最终教练组决定用两个办法传递帕切科的指挥：第一，帕切科把自己的想法写在小纸条上，再传递给教练组；第二，通过手机短信的形式发到路易斯的手机上。但考虑到信号可能不好，短信会有延迟，教练组还是决定以小纸条这种“鸡毛信”的方式作为第一方案，发短信作为第二方案。确定这两个方案之后，帕切科坐在哪里又成了一个大问题。如果是在工体，帕切科可以坐在主席台或者贵宾席上，但在客场，尤其是在西安这个客场，帕切科如果坐在看台上，甚至是主席台上，都有可能受到西安球迷的干扰。为了让帕切科有一个安静的环境，国安与陕西方面进行了反复沟通，最终把帕切科安排在体育场最高处几个平时用来转播的小房间之一。这个小房间也是现场转播的死角，因此在这场比赛中，不仅现场看不到帕切科坐在哪里，电视镜头也无法捕捉到他的身影，老帕就这样神秘“消失”了。比赛中，国安专门准备了两名工作人员传递鸡毛信，其中一人陪在帕切科身边。帕切科写完小纸条后，工作人员立刻坐电梯来到休息室，再由休息室外的副领队康玉明把小纸条传递给替补席上的路易斯。除此之外，帕切科还分别对场上每一个位置的对应换人进行了交代，这也是在张辛昕受伤情况不明时，替补席立刻招来小将姜涛准备替补的原因。比赛结束之后，帕切科也承认，自己不能在替补席指挥比赛，肯定会给球队造成一定影响。但这种经历自己过去也有过，他相信任何一名职业教练在执教生涯中都会遇到这种情况，所以自己也会有相应的办法应对。今天，路易斯和其他中方教练代替自己在替补席完成了指挥工作，自己很感谢他们。(宋词 发自西安)
\par\medskip
\textit{Deep Rewriting}\par\smallskip
帕切科遭禁赛无法用对讲机指挥 国安战术传达只能靠纸条 新浪体育讯 帕切科被停赛8场后，外界最关注的问题之一便是：无法坐上替补席的他，将如何指挥北京国安？尤其是在不能使用对讲机、手机等现代通讯手段直接沟通的情况下，帕切科的临场指挥方式更成了焦点。7月2日客场对阵陕西的比赛中，更令人意外的是，帕切科在赛前仿佛“消失”了一般。那么，他究竟是怎样带领球队完成这场艰苦比赛的呢？赛前联席会上，国安队领队魏克兴被明确告知，替补席上的教练员不得使用对讲机与外界沟通。事实上，这项规定早已有之，并非专门针对帕切科的停赛处罚，所有球队的替补席人员都不能使用对讲机。如此一来，国安此行带到西安的对讲机便无法派上用场，临场指挥的难度也随之增加。虽然足协并未限制使用手机，但考虑到比赛现场信号状况难以保证，加之看台环境嘈杂，通话效果并不稳定，国安教练组最终决定采用两套方案来传递帕切科的指令。第一种方式是由帕切科将自己的想法写在小纸条上，再传递给教练组；第二种方式则是通过手机短信发到助理教练路易斯的手机上。不过，由于担心短信可能因信号问题出现延迟，教练组最终将“小纸条”这种类似“鸡毛信”的方式确定为首选方案，短信则作为备用手段。确定传递方式后，帕切科坐在哪里又成了新的难题。如果是在工体，他可以坐在主席台或贵宾席上观察比赛；但在客场，尤其是在西安这样的环境中，帕切科若坐在看台甚至主席台上，都可能受到主队球迷的干扰。为了给他提供一个相对安静的指挥环境，国安与陕西方面进行了多次沟通，最终将帕切科安排在体育场最高处、平时用于转播的几个小房间之一。这个位置恰好也是现场转播的死角，因此整场比赛中，不仅现场观众看不到帕切科在哪里，电视转播镜头也无法捕捉到他的身影，老帕就这样神秘地“消失”了。比赛期间，国安专门安排了两名工作人员负责传递“鸡毛信”。其中一人始终陪在帕切科身边，等他写好纸条后，立刻乘电梯将指令送到休息室，再由等候在休息室外的副领队康玉明转交给替补席上的路易斯。除此之外，帕切科赛前还对场上各个位置可能出现的换人调整作了详细交代。这也解释了为何在张辛昕受伤情况尚不明确时，替补席能迅速叫来小将姜涛准备替补登场。赛后，帕切科承认，自己无法在替补席现场指挥，肯定会对球队造成一定影响。不过他也表示，类似经历自己过去曾经遇到过，相信任何一名职业教练在执教生涯中都可能面对这样的情况，因此他也有相应的应对办法。帕切科表示，本场比赛路易斯和其他中方教练代替自己在替补席完成了指挥工作，他对此非常感谢。（宋词 发自西安）
\par\medskip\hrule\medskip
\textbf{Prompt-based Generation}\par
\textit{Level 1}\par\smallskip
由于在停赛期间使用对讲机指挥比赛，帕切科再一次成为外界关注的焦点。昨天，中国足协有关人士明确表示，按照相关纪律规定，处于停赛期的教练员不得以任何方式进入替补席、技术区域，也不得通过通讯设备向场内传递指令。也就是说，在接下来的停赛场次中，帕切科如果想把自己的战术意图传达给球队，只能通过赛前布置以及比赛中由工作人员递送纸条等“原始方式”完成。在上一轮联赛中，因累积处罚无法坐上替补席的帕切科出现在看台包厢内。比赛进行过程中，电视镜头多次捕捉到他手持对讲机与场边人员交流。尽管俱乐部方面解释称，帕切科只是与教练组保持联系，并未直接干预比赛判罚或扰乱秩序，但这一做法仍引发争议。按照足协纪律准则，停赛并不只是“不能坐替补席”，更意味着教练员在比赛期间不得以现场指挥的形式继续履行主教练职责。足协相关部门在赛后对录像和比赛监督报告进行了核查，并向俱乐部进行了口头提醒。有关人士表示，停赛处罚的目的在于限制当事人在比赛中的指挥权，如果通过对讲机、电话等方式继续实时下达指令，实际上等同于规避处罚。“停赛教练可以观看比赛，但不能借助通讯工具指挥球队，这是原则问题。”该人士说。这也让帕切科接下来的临场安排变得更加困难。由于无法使用对讲机，教练组只能在赛前把可能出现的各种情况预先设计好，由临时负责临场指挥的助理教练根据比赛走势作出判断。如果看台上的帕切科发现需要调整，他可以通过工作人员递送纸条，将换人、阵型变化或注意事项写明后送到场边。不过，这种方式不仅效率低，而且存在信息延误，尤其是在比赛节奏较快、局势瞬息万变的情况下，很难做到及时反应。俱乐部方面对此显得比较谨慎。一位队内人士表示，球队会尊重足协规定，之后不会再让帕切科使用任何通讯设备指挥比赛。“主教练停赛确实会给球队带来影响，但规则必须遵守。我们会提前做好准备，让助理教练承担更多责任。”据了解，帕切科也已经在训练中加强了对替补席教练组的分工，重点强调比赛中几种常见局面的应对方案，尽量减少无法现场指挥带来的损失。对于球队而言，帕切科的激情和临场调度一直是重要特点。过去的比赛中，他经常在场边大声提醒队员站位，并根据对手变化迅速作出调整。如今被迫远离替补席，球队能否在缺少主教练实时指挥的情况下保持稳定，将成为下一场比赛的一大看点。球员们则表示，停赛不会改变球队的备战节奏，大家会按照教练组要求执行战术。事实上，教练停赛后通过各种方式“遥控”球队，在国内外赛场都并不鲜见。有的教练选择坐在看台上打电话，有的则安排专人传话，但随着竞赛管理越来越严格，这类做法正受到更明确限制。足协方面强调，今后如果再发现停赛人员使用通讯工具指挥比赛，不排除追加处罚的可能。帕切科能否适应只能靠纸条传达战术的特殊局面，将直接影响球队近期表现。对于他和俱乐部来说，如何在规则框架内完成指挥交接，或许比一场比赛的胜负更需要冷静处理。
\par\medskip
\textit{Level 2}\par\smallskip
帕切科停赛后，国安队的临场指挥方式成为外界关注的焦点。在无法像往常一样站在替补席前直接调度球队，也不能使用对讲机与替补席保持联系的情况下，这位葡萄牙主帅依然通过另一种方式参与了比赛指挥：小纸条和手机短信，成为他向场内传递战术意图的重要渠道。由于此前受到停赛处罚，帕切科本场比赛只能坐在看台上观战。按照相关规定，停赛教练不得进入替补席区域，也不能通过对讲机等即时通讯设备与场边工作人员直接交流。对于一支习惯由主教练在场边不断提醒、调整阵型和布置细节的球队来说，这无疑是一次特殊考验。比赛开始后，国安替补席上承担主要指挥职责的是助理教练路易斯以及其他中方教练组成员，他们需要在场边完成换人、战术提醒和情绪管理等工作。从比赛现场情况看，帕切科虽然不在替补席，但并未完全“离开”指挥体系。坐在看台上的他始终密切关注场上局势，不时与身边工作人员交流。每当他认为需要对场上站位、攻防节奏或人员安排进行提醒时，便会将自己的意见写在纸条上，或通过手机短信传递给相关工作人员，再由工作人员转达给替补席。随后，路易斯和中方教练根据场上实际情况，将这些信息转化为具体指令，传递给球员。这种“隔空指挥”并不如场边直接喊话那样迅速，但在特殊规则限制下，已经成为教练组能够采取的有效办法。为了避免信息传递出现延误，国安教练组赛前就进行了预案准备，对比赛中可能出现的不同局面作了分工。路易斯主要负责临场战术执行，中方教练则更多承担与球员沟通、提醒站位以及协助换人的工作。帕切科提供的意见，更多是从整体节奏和战术方向上给予提示。据现场观察，国安替补席在比赛中并未因主教练缺席而显得混乱。路易斯多次走到技术区域内向队员做出手势，提醒中后场保持距离，也在进攻推进受阻时要求边路加快出球速度。中方教练则不断在替补席前与热身球员交流，准备随时根据比赛需要进行人员调整。几次关键换人安排，均由替补席教练组完成执行，而帕切科的意见也通过纸条或短信的方式被及时纳入讨论。对于国安来说，帕切科停赛带来的影响不仅体现在临场指挥方式的变化上，也考验着整个教练团队的协作能力。过去，帕切科在场边以情绪饱满、指挥频繁著称，他的临场反应往往直接影响球队节奏。如今他被迫退居看台，助理教练和中方教练必须承担更多责任，既要贯彻主教练赛前布置，又要根据场上变化独立作出判断。赛后，教练组内部普遍认为，本场指挥工作总体运行顺畅。虽然沟通方式受到限制，但赛前准备较为充分，替补席与看台之间的信息传递基本保持连贯。路易斯和其他中方教练完成了帕切科缺席情况下的场边指挥任务，也让球队在特殊环境下维持了既定战术框架。帕切科的停赛仍将持续一段时间，国安接下来还可能继续面对类似情况。如何在规则允许范围内保持教练组运转效率，如何让替补席在没有主教练直接坐镇的情况下做出更快速判断，将是球队必须继续解决的问题。至少从目前情况看，小纸条和手机短信虽然不是最理想的指挥方式，却帮助帕切科在看台上延续了对球队的影响，而路易斯和中方教练的执行，也让国安度过了停赛后的首个临场指挥考验。
\par\medskip
\textit{Level 3}\par\smallskip
7月2日，在国安客场对阵陕西的比赛中，被停赛8场的帕切科无法坐上替补席指挥，也不能通过对讲机与替补席沟通。为保证临场指令传递，国安教练组赛前制定了“小纸条为主、手机短信为辅”的方案，最终由路易斯和其他中方教练在替补席完成现场指挥。这场比赛的地点在西安。此前，帕切科因受到停赛处罚，无法按照以往方式在替补席直接指挥球队。与此同时，替补席教练员不得使用对讲机与外界沟通的要求，也在赛前被再次明确。国安方面了解到，这一规定并非专门针对帕切科的停赛而设，而是足协此前早已作出的规定，比赛中替补席人员不得通过对讲机同看台或其他区域进行联系。据悉，在比赛前的联席会上，国安方面的魏克兴被告知，替补席上的教练员不能使用对讲机与外界沟通。由于帕切科无法出现在替补席，国安教练组必须寻找其他指挥方式。经过沟通和研究，教练组决定将传递小纸条作为第一方案，将手机短信作为第二方案，以便帕切科在观看比赛后及时把自己的判断和调整意见传递到替补席。为了让帕切科能够尽可能清楚地观察比赛，国安与陕西方面进行了沟通。随后，帕切科被安排在体育场最高处用于转播的小房间之一。这个位置虽然远离替补席，但视野较高，有利于他观察比赛整体走势、队形变化以及各个位置的运行情况。对国安来说，在不能进行直接临场指挥的情况下，选择一个便于判断场上局面的观察点，是保障指挥信息有效传递的前提。比赛过程中，国安按照赛前方案安排工作人员负责传递纸条。帕切科在小房间内观察比赛后，将自己的意见通过身边人员传送到休息室，再由康玉明把信息转交给替补席上的路易斯。路易斯收到指令后，再结合替补席上中方教练的判断，对场上队员作出相应安排。手机短信则作为备用方式，在纸条传递不便或需要补充说明时使用。帕切科赛前已经对场上每个位置可能出现的对应换人作了交待，包括不同位置出现体能、战术或人员调整需要时，替补席应如何应对。张辛昕、姜涛等队员所在位置涉及的调整，也都被纳入预案范围。这样一来，即使帕切科不能在场边直接喊话，替补席仍能根据既定预案和临场传来的信息执行调整。在具体执行中，路易斯成为替补席上接收并落实指令的重要人物。由于帕切科不能坐在教练席，他需要依靠路易斯以及其他中方教练完成现场沟通。魏克兴、康玉明等人则在不同环节中承担协调和传递作用，确保从看台高处到替补席之间的信息链条保持通畅。从比赛组织角度看，国安此次指挥方式是在规则限制下作出的应对。足协关于替补席不得使用对讲机与外界沟通的规定，使球队不能采用最直接的远程联系手段；而停赛处罚又使帕切科无法在替补席亲自指挥。两方面因素叠加，促使国安必须在赛前完成更细致的预案，并在比赛中通过人工传递纸条等方式完成临场沟通。最终，路易斯和其他中方教练在替补席上代替帕切科完成了指挥工作。虽然这种方式无法完全等同于主教练现场坐镇，但国安通过赛前安排、人员分工和信息传递，尽量降低停赛带来的影响。帕切科也承认，自己不能在替补席指挥比赛，肯定会给球队造成一定影响，但他表示自己会有相应办法应对，并感谢路易斯和其他中方教练在比赛中的配合与工作。新浪体育讯，宋词发自西安。
\par\medskip
\textit{Level 3+Ref.}\par\smallskip
7月2日，在西安进行的国安与陕西比赛中，因被停赛而无法坐上替补席的帕切科，仍通过事先准备的方式参与球队临场指挥。在替补席教练员不得使用对讲机与外界沟通的规定下，国安教练组最终以小纸条作为第一传递方案、手机短信作为第二方案，将帕切科的指挥信息送到替补席，由路易斯和其他中方教练完成现场执行。帕切科此前被停赛8场，这意味着他不能在比赛中进入替补席指挥国安。围绕停赛后的临场指挥问题，国安在赛前就开始准备相应办法。7月2日赛前联席会上，国安方面被明确告知，替补席上的教练员不得使用对讲机与外界沟通。魏克兴在联席会上获知这一要求后，也将情况反馈给教练组。需要说明的是，替补席人员不得使用对讲机并非针对帕切科本次停赛临时作出的安排，而是此前早已存在的相关规定。在无法通过对讲机直接沟通的情况下，国安教练组很快确定了应对方案。按照安排，帕切科的指挥意见首先通过小纸条传递；如果小纸条传递不便，则以手机短信作为第二选择。这样的安排，是为了尽可能保证主教练对比赛形势的判断能够及时传达到替补席，同时也符合比赛管理规定。为了让帕切科能够更清楚地观察比赛，国安方面赛前还与陕西方面进行了沟通。经过协调，帕切科被安排在体育场最高处、用于电视转播的小房间之一。这个位置可以俯瞰比赛场地，便于他观察双方阵型变化、球员跑位以及各个位置上的临场情况。虽然无法像平时一样在替补席近距离指挥，但这一位置至少为他进行全局判断提供了条件。比赛期间，国安按照赛前确定的流程执行传递工作。俱乐部安排工作人员负责传递小纸条，帕切科身边人员将写有指挥信息的纸条送到休息室，再由康玉明把信息传给替补席上的路易斯。路易斯随后结合替补席现场情况，与其他中方教练一起将相关安排传达到场内球员。整个过程虽然比直接口头指挥更复杂，但国安希望通过这种方式尽量减少帕切科停赛对球队临场运转造成的影响。据悉，帕切科在赛前已对场上每个位置可能出现的对应换人进行了交待。也就是说，在比赛开始前，他已经把不同情况下的人员调整方案告知教练组。这样一来，即使比赛中无法直接站在场边喊话，替补席上的教练仍可根据预案和帕切科传递来的即时判断作出处理。对于一名被停赛的主教练而言，这种预先部署加临场传递的方式，是国安在现有条件下采取的主要应对办法。在实际指挥中，路易斯和其他中方教练承担了帕切科原本在替补席上的职责。他们不仅要关注场上局势，还要接收和理解来自看台高处的指令，再将这些信息转化为场边指挥。魏克兴、康玉明等教练组成员也都参与到这一沟通链条之中，确保信息从帕切科所在位置传递到替补席，再由替补席落实到比赛当中。帕切科也承认，自己不能在替补席指挥比赛，肯定会给球队造成一定影响。与正常情况下相比，他无法第一时间与球员沟通，也不能直接通过临场动作和喊话调整球队。不过他同时表示，自己会有相应办法应对这一困难，并对路易斯和其他中方教练在比赛中的配合表示感谢。从此次比赛的组织过程看，帕切科停赛后的指挥方式并非单一依赖某一种手段，而是国安教练组在规则限制下作出的系统安排。小纸条和手机短信成为替代对讲机的沟通渠道，高处转播房间成为帕切科观察比赛的位置，路易斯和中方教练则在替补席完成最终执行。相关消息由新浪体育讯报道，宋词发自西安。
\end{promptbox}
\captionof{figure}{A paired example from the News domain, comprising one human-written source and eleven machine-involved variants.}
\label{fig:paired-data-example}
\end{CJK*}

\end{document}